\documentclass[10pt,twocolumn,letterpaper]{article}

\usepackage{wacv}      %

\usepackage{graphicx}
\usepackage{amsmath}
\usepackage{amssymb}
\usepackage{booktabs}
\usepackage{xcolor}
\definecolor{darkblue}{rgb}{0, 0, 0.5}
\usepackage{microtype}
\usepackage{placeins}
\usepackage{fontawesome5}
\usepackage{algorithm}
\usepackage{algorithmic}
\usepackage{times}
\usepackage{epsfig}
\usepackage{graphicx}
\usepackage{xcolor}
\usepackage{amsmath}
\usepackage{amsfonts}
\usepackage{makecell}
\usepackage{algorithm}
\usepackage{algorithmic}
\usepackage{float}
\usepackage{tabularx}
\usepackage{longtable}
\usepackage{supertabular}
\usepackage{xltabular}
\usepackage{array}
\usepackage{colortbl}
\usepackage{adjustbox}
\usepackage{soul}
\usepackage{newfloat}
\usepackage{listings}
\usepackage{placeins}
\usepackage{caption}
\usepackage{subcaption}
\usepackage{adjustbox}
\usepackage{tikzsymbols}
\usepackage{amsmath}
\usepackage{booktabs}

\usepackage{multirow}
\usepackage[skins]{tcolorbox} %

\definecolor{valbest}{HTML}{d9ead3}
\newcommand{\valbest}[1]{\colorbox{valbest}{#1}}
\definecolor{valgood}{HTML}{cfe6ec}
\newcommand{\valgood}[1]{\colorbox{valgood}{#1}}
\definecolor{valmid}{HTML}{fce5cd}

\definecolor{valbad}{HTML}{ead1dc}

\definecolor{themegreen}{HTML}{365956}
\definecolor{themepurple}{HTML}{3c1b48}
\definecolor{themered}{HTML}{b43748}

\usepackage{enumitem}
\newlist{noindentenumerate}{enumerate}{1}
\setlist[noindentenumerate]{
  label=\arabic*., %
  left=0pt, %
  itemsep=-1ex, %
}
\everypar{\looseness=-1}

\newcommand{\iiitdlogo}{\raisebox{5pt}{\includegraphics[scale=0.0113]{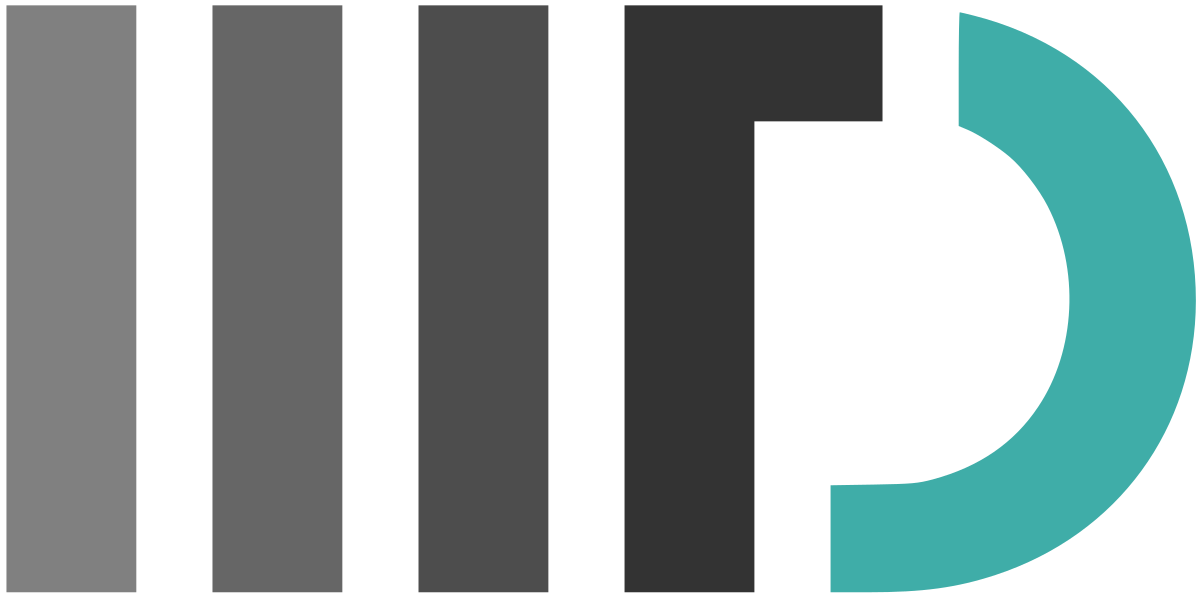}}}
\newcommand{\ublogo}{\raisebox{5.2pt}
{\includegraphics[scale=0.085]{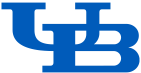}}}

\newcommand{\adobelogo}{\raisebox{5pt}{\includegraphics[scale=0.032]{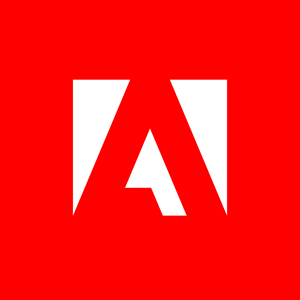}}}
\newcommand{\bitslogo}{\raisebox{5pt}{\includegraphics[scale=0.009]{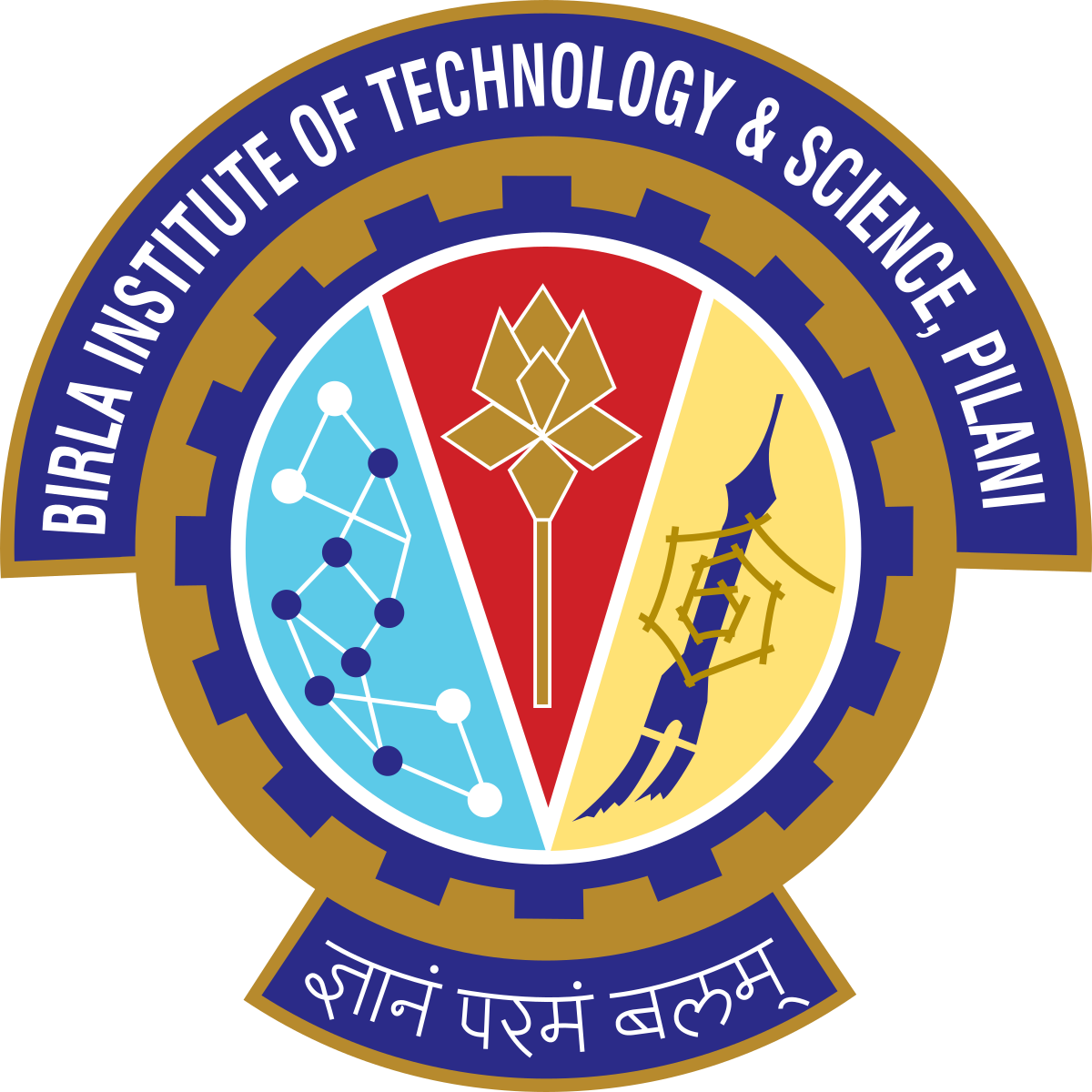}}}
\newcommand\coauth{$^\star$}
\newcommand\blfootnote[1]{%
  \begingroup
  \renewcommand\thefootnote{}\footnote{#1}%
  \addtocounter{footnote}{-1}%
  \endgroup
}

\usepackage[pagebackref,breaklinks,colorlinks]{hyperref}

\usepackage[capitalize]{cleveref}
\crefname{section}{Sec.}{Secs.}
\Crefname{section}{Section}{Sections}
\Crefname{table}{Table}{Tables}
\crefname{table}{Tab.}{Tabs.}

\def\wacvPaperID{1596} %
\def\confName{WACV}
\def\confYear{2025}

\begin{document}

\title{\vspace*{-0.35in} Long-Term Ad Memorability: Understanding \& Generating Memorable Ads}

\author {
    \textbf{Harini SI}\coauth \adobelogo \hspace{0.8em}
    \textbf{Somesh Singh}\coauth \adobelogo \hspace{0.8em}
    \textbf{Yaman K Singla}\coauth \adobelogo  \hspace{0.1mm} \ublogo \hspace{0.1mm} \iiitdlogo \hspace{0.3mm} \textbf{Aanisha Bhattacharyya}\adobelogo \hspace{0.8em}\\
    \textbf{Veeky Baths}\bitslogo \hspace{0.8em}
    \textbf{Changyou Chen}\ublogo \hspace{0.8em}
    \textbf{Rajiv Ratn Shah}\iiitdlogo \hspace{0.8em}
    \textbf{Balaji Krishnamurthy}\adobelogo \hspace{0.8em}\\
    {\small \adobelogo Adobe Media and Data Science Research, \ublogo State University of New York at Buffalo, \iiitdlogo IIIT-Delhi, \bitslogo BITS Pilani, Goa}\\
    {\href{mailto:behavior-in-the-wild@googlegroups.com}{behavior-in-the-wild@googlegroups.com}}
}

\maketitle

\begin{abstract}
 Despite the importance of long-term memory in marketing and brand building, until now, there has been no large-scale study on the memorability of ads. All previous memorability studies have been conducted on short-term recall on specific content types like action videos. On the other hand, long-term memorability is crucial for advertising industry, and ads are almost always highly multimodal. Therefore, we release the first memorability dataset, LAMBDA, consisting of 1749 participants and 2205 ads covering 276 brands. Running statistical tests over different participant subpopulations and ad types, we find many interesting insights into what makes an ad memorable, \textit{e.g.}, fast-moving ads are more memorable than those with slower scenes; people who use ad-blockers remember a lower number of ads than those who don't. Next, we present a model, Henry, to predict the memorability of a content. Henry achieves state-of-the-art performance across \textit{all} prominent literature memorability datasets. It shows strong generalization performance with better results in 0-shot on unseen datasets. Finally, with the intent of memorable ad generation, we present a scalable method to build a high-quality memorable ad generation model by leveraging automatically annotated data. Our approach, SEED (Self rEwarding mEmorability Modeling), starts with a language model trained on LAMBDA as seed data and progressively trains an LLM to generate more memorable ads. We show that the generated advertisements have 44\% higher memorability scores than the original ads. We release this large-scale ad dataset, UltraLAMBDA, consisting of 5 million ads. Our code and the datasets, LAMBDA and UltraLAMBDA, are open-sourced at \url{https://behavior-in-the-wild.github.io/memorability}. %
\end{abstract}

\blfootnote{$^*$Equal Contribution. Contact \href{mailto:behavior-in-the-wild@googlegroups.com}{behavior-in-the-wild@googlegroups.com} for questions and suggestions}

\vspace{-0.15in}

\section{Introduction}
\label{sec:introduction}
\textit{``The first lesson of branding: memorability. It is very difficult buying something you can't remember.''} - Sir John Hegarty, the creator of the iconic ads for Levi's, Nike, Microsoft, Tinder, and Coke.

The global advertising industry is \$700 billion+ industry \cite{forbes2021}. Three out of the ten largest companies by market capitalization are advertising companies with average revenues exceeding \$250 billion. The World Wide Web is mostly funded by advertising. Given that marketers are spending such large sums of money on advertisements, it is imperative to know if their brand would even be recalled at the customer's purchase time. This would help the marketers optimize their costs, content, delivery, and audience, ultimately helping in boosting sales. Most of the studies carried out in the machine learning literature have been on short-term memorability (memorability testing in less than 5 minutes) on action videos like walking and dancing (Table~\ref{tab:dataset_comparison}). On the other hand, customer purchase decisions are rarely carried out within five minutes of watching an ad. In fact, the marketing funnel model popular in the marketing literature says that customers pass through several stages of a funnel, like awareness and consideration, before the actual sale \cite{lavidge1961model}. Further, in the ML literature, there have been no memorability studies on advertisements. Advertisements are highly multimodal; they contain video, speech, music, text overlaid on scenes, jingles, specific brand colors, \textit{etc}. None of these elements are found in previous studies like VideoMem, Memento10k, LaMem, \textit{etc.} (refer to Table~\ref{tab:dataset_comparison} for a detailed comparison).

\begin{table*}[!th]
\vspace*{-0.3in}
\centering

\resizebox{1.0\textwidth}{!}{
\scriptsize
\begin{tabularx}{1.0\textwidth}{>{\hsize=0.4\hsize\bfseries}X>{\hsize=0.3\hsize}X>{\hsize=0.5\hsize}X>{\hsize=0.35\hsize}X>{\hsize=0.85\hsize}X>{\hsize=0.35\hsize}X>{\hsize=0.3\hsize}X>{\hsize=0.3\hsize}X>{\hsize=0.4\hsize}X}
\hline
\textbf{Dataset} & \textbf{\#Samples} & \textbf{Memory Type} & \textbf{Memory Retrieval Process} & \textbf{Content} & \textbf{Average Screen Duration} & \textbf{Audio Present} & \textbf{Human Consistency} & \textbf{Memorability Measurement Protocol} \\ \hline

\makecell{\textbf{Memento10k}} & 10,000 & ST ($<$ 10 mins) & Recognition & Videos of single type of action obtained from amateur videos & 3s & Yes & 0.73 & Competition \\

\makecell{\textbf{VideoMem}} & 10,000 & ST (few mins), LT (1-3 days) & Recognition & Videos of a single type of action obtained from professional (staged) footage & 7s & None & 0.48 (ST), 0.19 (LT) & Competition\\

\makecell{\textbf{LaMem}} & 60,000 & ST ($<$ 3.5 mins) & Recognition & General Images & 0.6s & None & 0.68 & Competition\\

\makecell{\textbf{SUN}} & 2,222 & ST ($<$ 4.4 mins) & Recognition & General Images & 1s & None & 0.75 & Competition \\

\makecell{\textbf{MemCat}} & 10,000 & ST ($<$ 3.5 mins) & Recognition & General Images & 0.6s & None & 0.78 & Competition \\

\makecell{\textbf{MediaEval}} & 1500 & ST (few mins) and LT ($<$ 3 days) & Recognition & Short video clips collected from Twitter and Flickr & 6s & None & - & Competition\\

\textbf{LAMBDA (Ours)} & 2,205 & LT (1-3 days) & \textbf{Recognition and Recall }& Videos of multimodal advertisements & \textbf{33s} & \textbf{Yes} & \textbf{0.61} & \textbf{Natural} \\ \hline
\end{tabularx}%
}
\caption{\label{tab:dataset_comparison} Comparison of all the major (image and video) memorability datasets available in the literature along with LAMBDA (ours). The datasets are compared on the following axes: number of samples, type of memorability (short-term (ST) and long-term (LT)), memory retrieval process (recall or recognition), type of content (images/videos and their type), duration with which the sample was shown on the participants' screen, whether audio was present or not, human consistency achieved in the study, and the protocol followed in the study to collect the data.
\textbf{Memento10k} - \protect\cite{newman2020multimodal},
\textbf{VideoMem} - \protect\cite{cohendet2019videomem},
\textbf{LaMem} - \protect\cite{khosla2015understanding},
\textbf{SUN} - \protect\cite{isola2011makes},
\textbf{MemCat} - \protect\cite{goetschalckx2019memcat},
\textbf{MediaEval} - \protect\cite{Kiziltepe_2021}
}
\vspace*{-0.1in}
\end{table*}

\noindent \textbf{What drives memory?} Memory over content is determined by two factors: human factors and the content itself \cite{bylinskii2015intrinsic}. Human factors represent the viewer's thoughts, emotions, and actions, while the content factors are words and raw pixels of text, images, and videos. Foundational large-scale studies on memorability \cite{isola2011makes,khosla2015understanding,cohendet2019videomem,akagunduz2019defining} showed that there is sufficient consistency between humans in what they remember. Human-human memorability consistency scores are in the range of 0.6-0.8. This means that the memorability ranks of a content between two groups of humans are more than 60\% correlated.

These initial studies also tried to answer the question of what makes a content memorable. They found that low-level image features like colors, aesthetics, number of objects, and such have very little correlation with whether the image was remembered. On the other hand, high-level features like object and scene semantics have significant correlation with memorability. For example, human images are more memorable than object images. %
Further, these initial studies contributed to protocols for conducting memorability studies. They proposed a competitive memorability game, where they asked participants to recognize as many images as they could remember. The game ended for those participants whose scores fell below certain success rate thresholds. However, this protocol limits the scope of these studies to short-term memorability (a few seconds to a few minutes), and the competitive nature makes the study unnatural and, thus, not applicable to real-world scenarios like marketing where the customers are not competing with each other to remember the brand. Therefore participants in all these studies are aware that they are being tested for memorability, this can create a deviation from their natural behaviour commonly known as the Hawthorne effect in psychology \cite{mccarney2007hawthorne,roethlisberger2003management,malavolta2022awareness}

\noindent\textbf{What drives customer memory?} Customer purchase decision is a long process. Marketing theory formulates this as a funnel where customers pass through several stages like awareness, consideration, and evaluation before the actual sale \cite{lavidge1961model}. Due to the purchase funnel being a multi-stage lengthy process, long-term memorability (LTM) is the closest proxy to model customer memory \cite{norris2017short,waugh1965primary}. While the LTM store (as distinct from the STM store) has been studied for over five decades in psychology \cite{ebbinghaus1885gedachtnis,atkinson1968human}, there are no large-scale studies containing data over such time period that can help us model the long-term customer LTM spanning days or more \cite{norris2017short,waugh1965primary}. Unfortunately, STM datasets, typically measuring memorability of a few seconds to a few minutes, are not good proxies to model customer memory. Moreover, the competitive nature of the memorability games in the previous studies further disconnect the modeling from advertising use cases.

\begin{figure*}[]
    \centering
    \vspace*{-0.25in}
    \begin{subfigure}{0.18\textwidth}
        \centering
        \includegraphics[width=\textwidth]{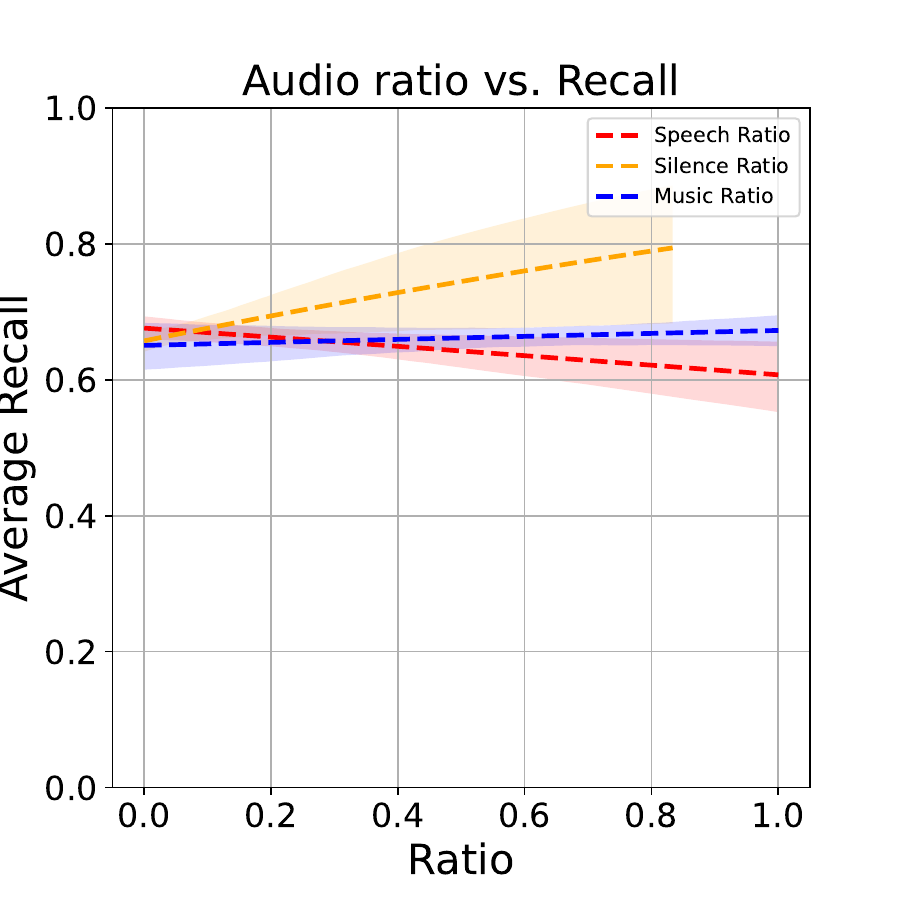}
        \caption{}
        \label{subfig:speech vs recall}
    \end{subfigure}
    \begin{subfigure}{0.18\textwidth}
        \centering
        \includegraphics[width=\textwidth]{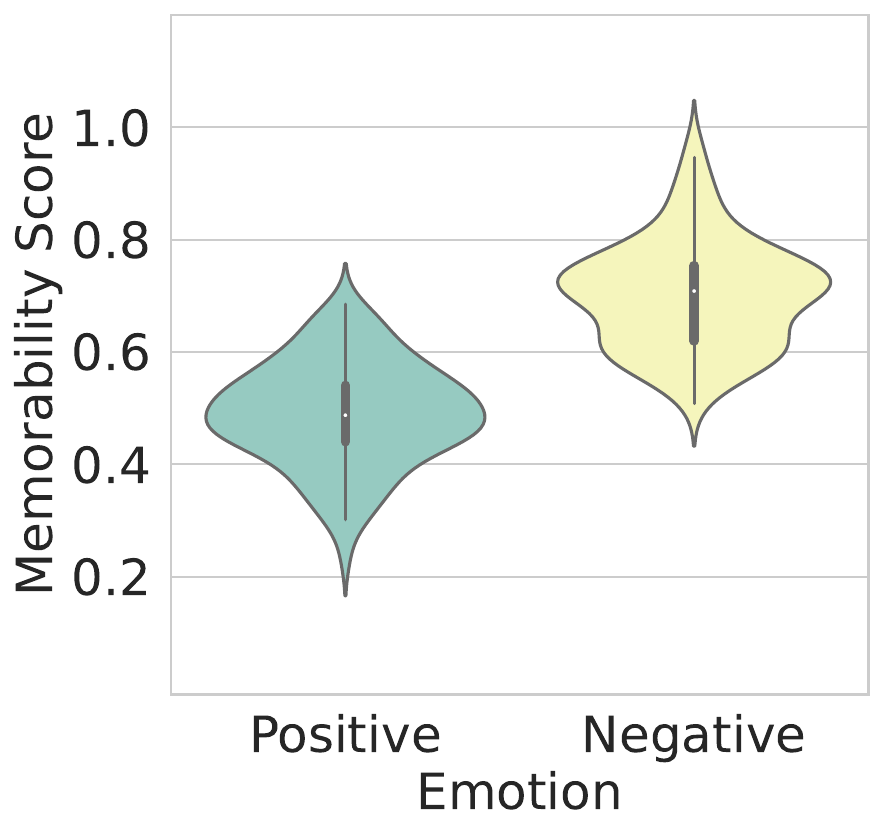}
        \caption{}
        \label{subfig:emotion vs recall}
    \end{subfigure}
    \begin{subfigure}{0.18\textwidth}
        \centering
        \includegraphics[width=\textwidth]{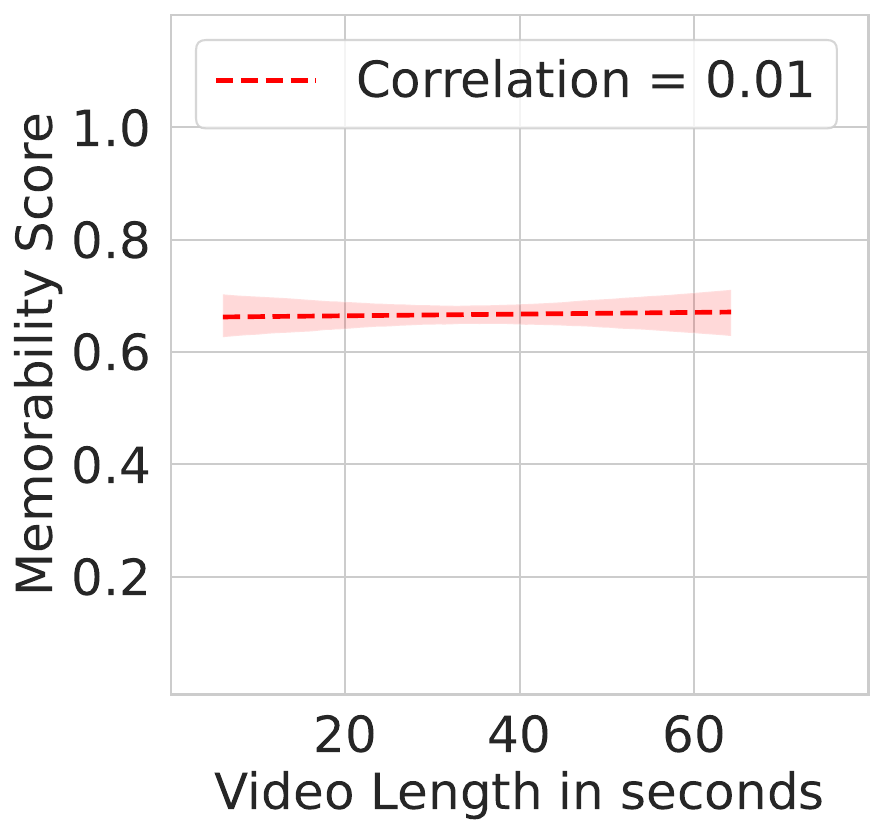}
        \caption{}
        \label{subfig:length vs mem}
    \end{subfigure}
    \begin{subfigure}{0.18\textwidth}
        \centering
        \includegraphics[width=\textwidth]{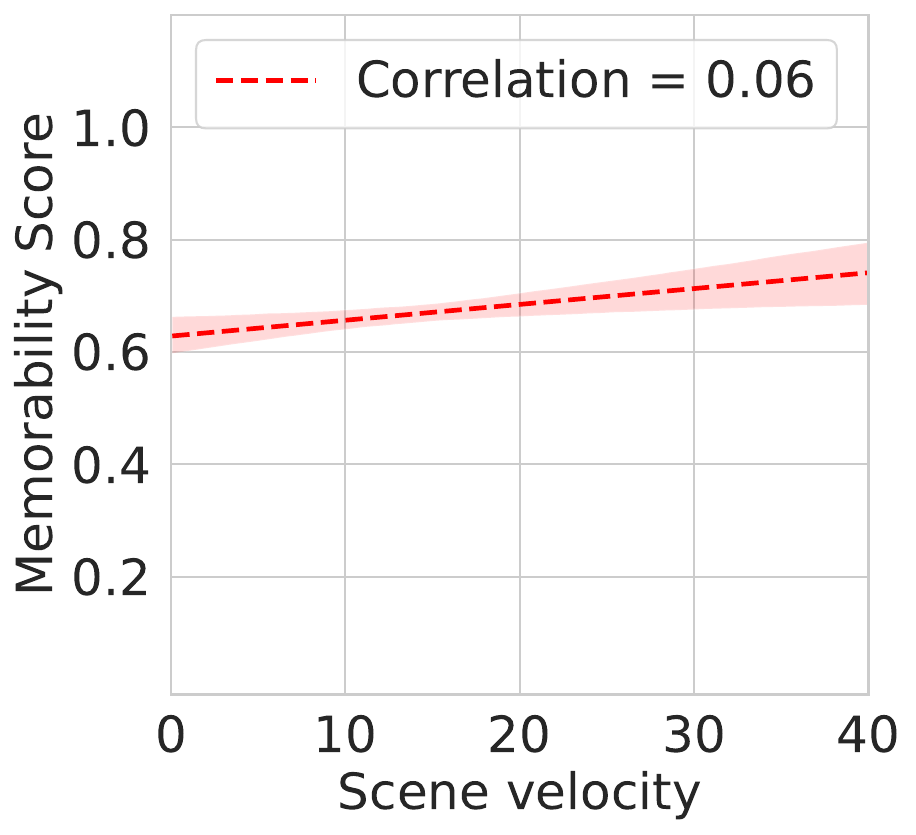}
        \caption{}
        \label{subfig:velocity vs recall}
    \end{subfigure}
    \begin{subfigure}{0.18\textwidth}
        \centering
        \includegraphics[width=\textwidth]{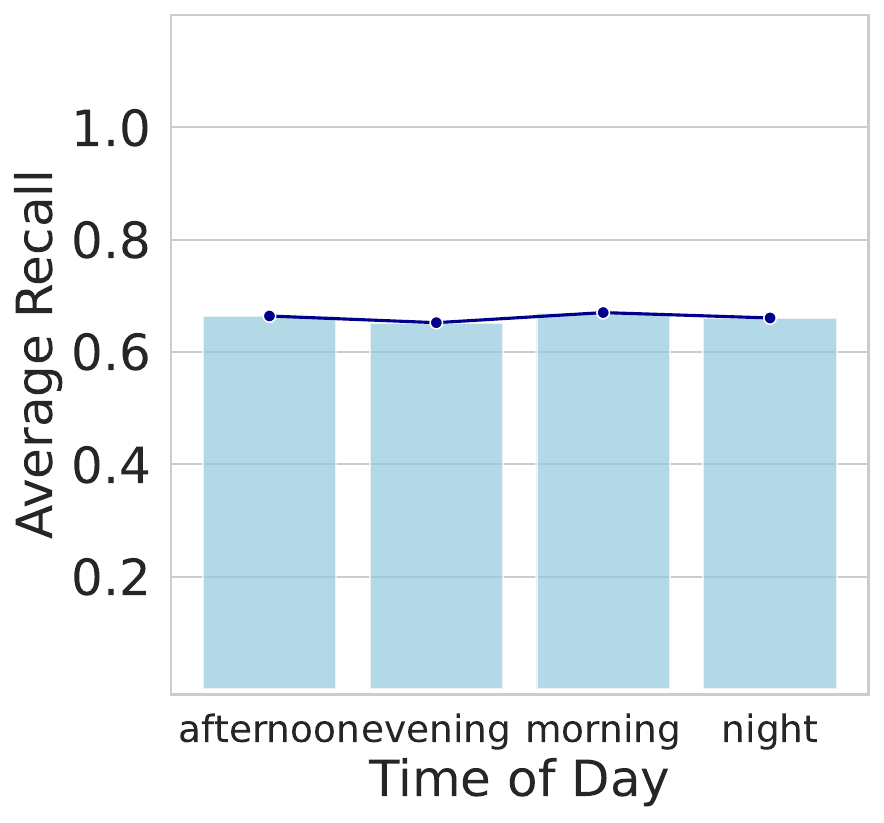}
        \caption{}
        \label{subfig:avg recall by time}
    \end{subfigure}
      \begin{subfigure}{0.18\textwidth}
        \centering
        \includegraphics[width=\textwidth]{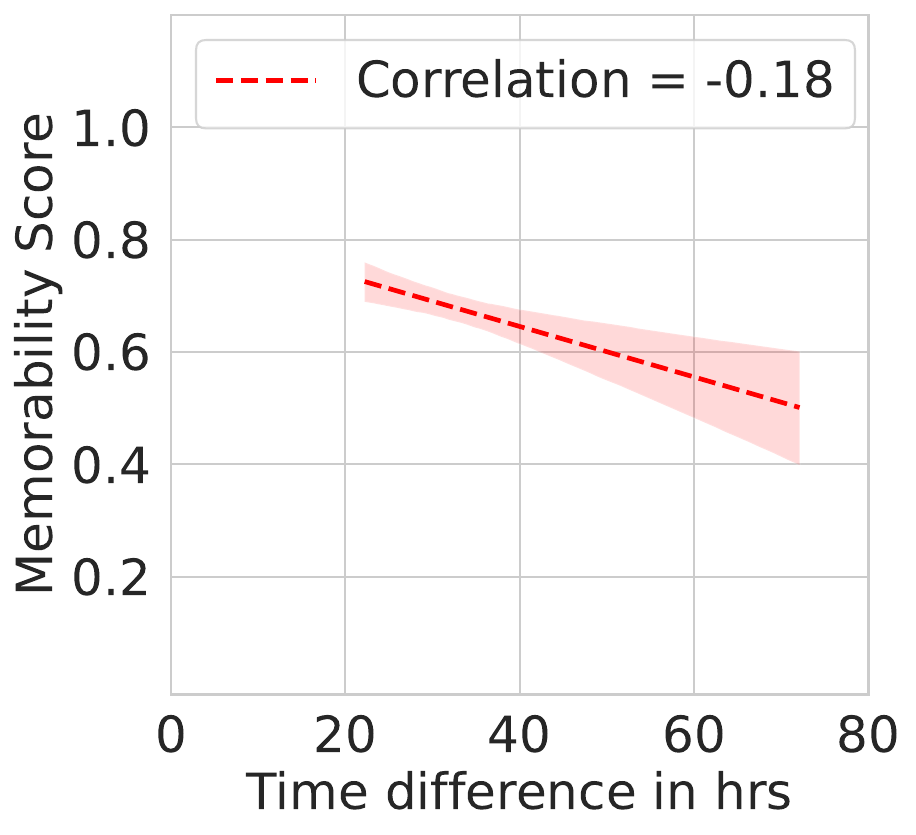}
        \caption{}
        \label{subfig:delay vs recall}
    \end{subfigure}
    \begin{subfigure}{0.18\textwidth}
        \centering
        \includegraphics[width=\textwidth]{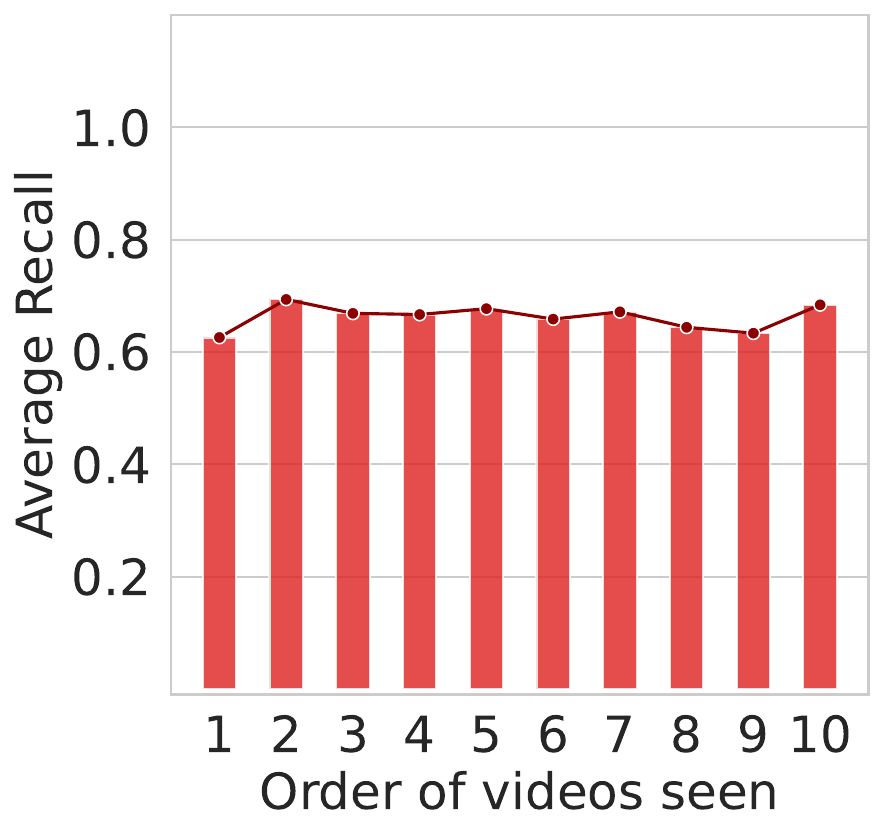}
        \caption{}
        \label{subfig:avg recall by position}
    \end{subfigure}
    \hfill
    \begin{subfigure}{0.18\textwidth}
        \centering
        \includegraphics[width=\textwidth]{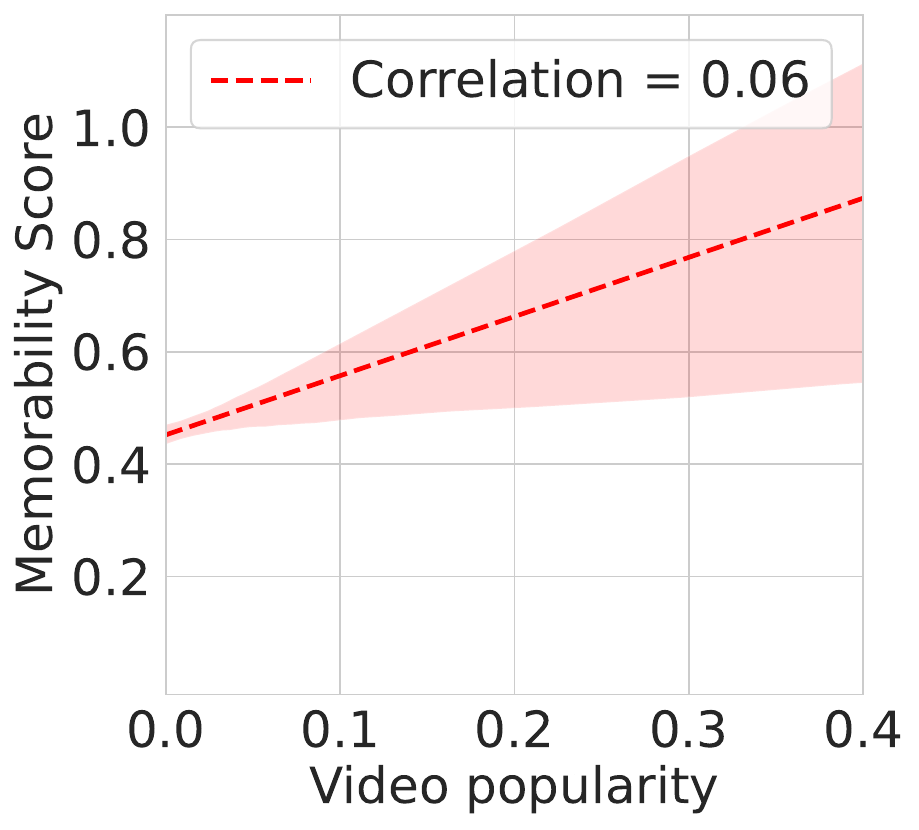}
        \caption{}
        \label{subfig:popularity vs recall}
    \end{subfigure}
    \hfill
    \begin{subfigure}{0.18\textwidth}
        \centering
        \includegraphics[width=\textwidth]{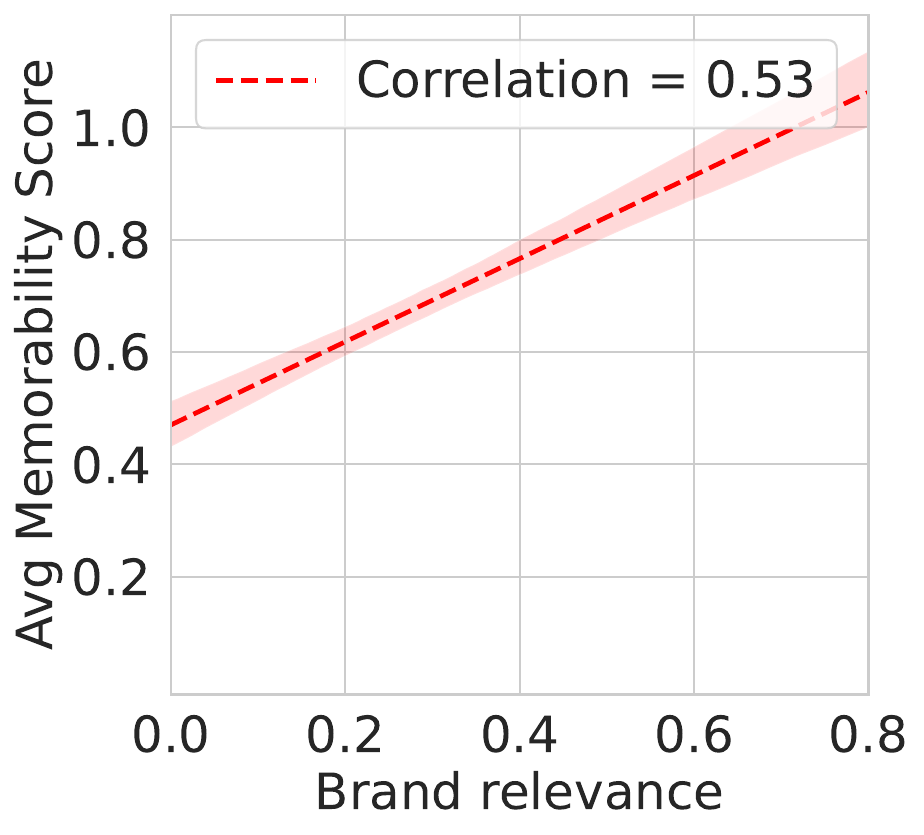}
        \caption{}
        \label{subfig:relevance vs recall}
    \end{subfigure}
    \hfill
    \begin{subfigure}{0.18\textwidth}
        \centering
        \includegraphics[width=\textwidth]{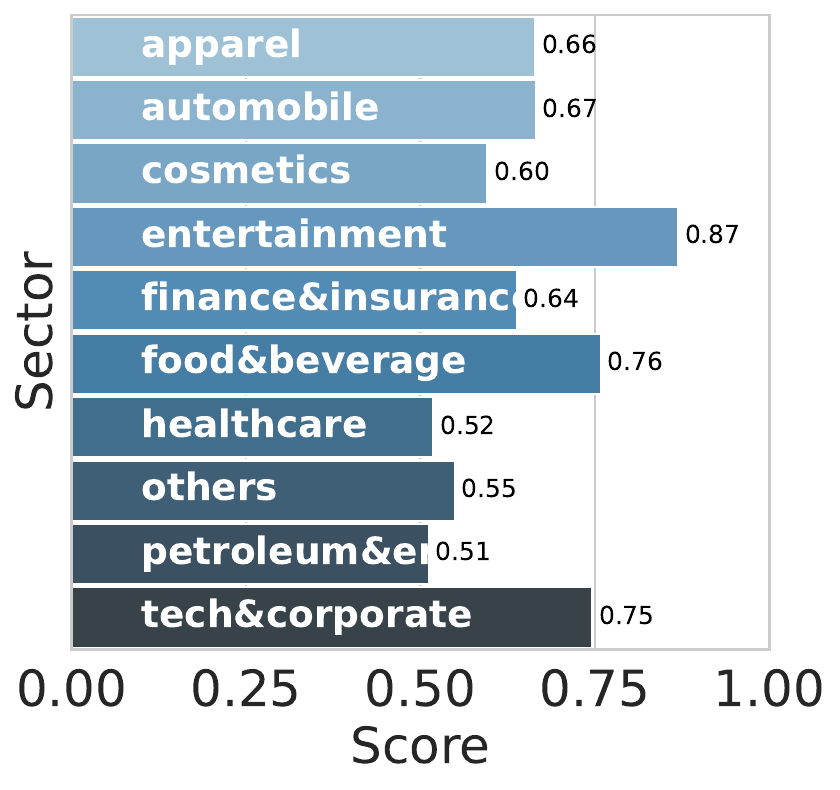}
        \caption{}
        \label{subfig:recall by sector}
    \end{subfigure}
    \hfill
    \caption{Correlations between \textit{content factors} (a-d), \textit{interaction factors} (e-g), and \textit{customer behavior factors} (h-j) with memorability on LAMBDA samples. While emotion has a high correlation with memory, other content factors do not have much correlation. Further, while there is little correlation between the order of videos seen and memorability; with time, participants' memory of the videos shows a forgetting trend. Video popularity, as measured by YouTube likes/views, shows a slight positive correlation with memory. Average brand relevance has a strong positive correlation with memory, with top sectors being remembered as food, entertainment, and tech. Speech, silence and music have little effect with silence having the highest positive correlation with recall. Silence ratio is measured as the percentage of silence in a video, similarly for music and speech.}
    \label{fig:correlation-graphs}
    \vspace*{-3mm}
\end{figure*}

\begin{figure}[]
    \centering
    \includegraphics[width=0.45\textwidth]{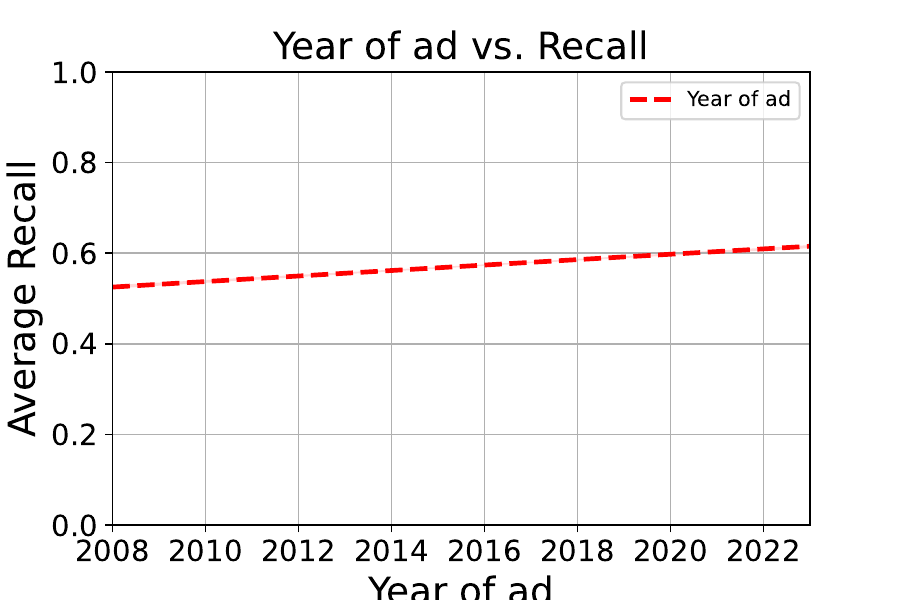}
    \label{fig:memorabilityvides}

    \caption{The graph shows the relationship of the year the ad is uploaded on youtube vs the recall. 
    }
    \label{fig:Year}
\end{figure}

To answer the question of what drives customer memory, there have been efforts in marketing literature where researchers have conducted many field experiments with the intent to prove certain hypotheses. For instance, Li \textit{et al.} \cite{li2010primacy} conducted a field experiment on advertisements shown during the 2006 Super Bowl Games where they asked the audience to recall the brands they saw in the game held (at least) a day earlier. They found a strong primacy effect, where viewers remembered brands more if they occurred earlier when controlling for the commercial length. Similarly, there have been studies to determine the effect of syntactic complexity \cite{atalay2023creating}, emotional content \cite{putrevu2004consumer,mai2009emotions}, repetition \cite{schmidt2015advertising}, spot length \cite{newstead2010cost,varan2020effects}, the position of brand logo and imagery \cite{newstead2010cost}, sound presence \cite{bellman2021can}, and on customer factors like involvement and relevance \cite{newstead2010cost,schmidt2015advertising}.

\begin{figure*}[!t]
\vspace*{-0.3in}
    \centering
    \includegraphics[width=0.85\textwidth]{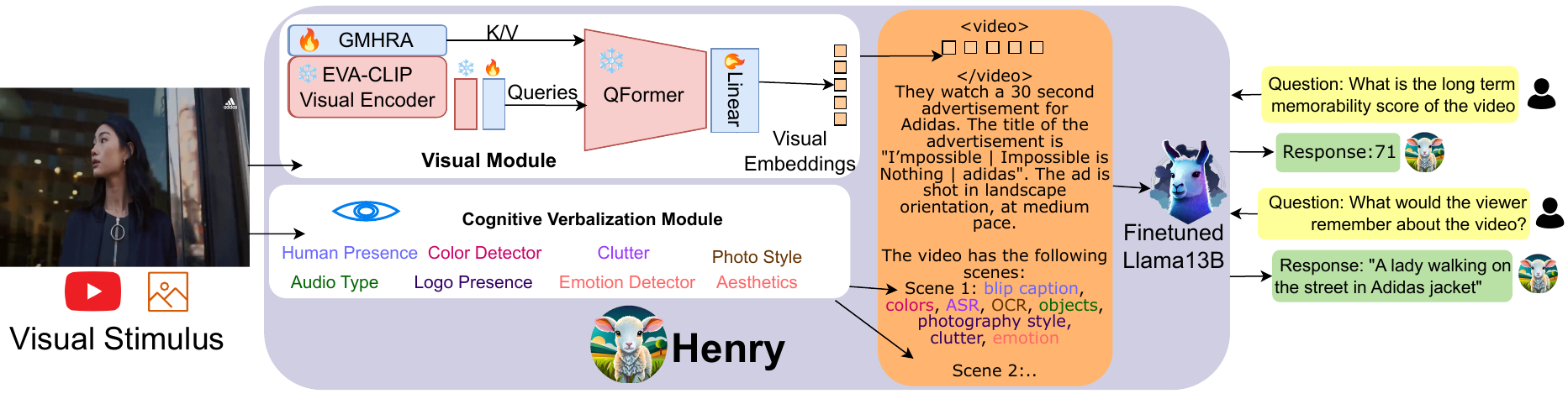}
    \caption{Predicting memorability by encoding visual information (via visual encoder EVA-CLIP), cognitive concepts (via verbalization module), and world knowledge (through fine-tuned Llama). We instruction fine-tune the combined model end to end to predict user memorability. Snowflake and fire symbols denote the frozen and unfrozen parts of the architecture.}
    \label{fig:memorability-model}
\vspace*{-4mm}
\end{figure*}

While these studies have contributed much towards understanding the factors that drive customer memory, they are limited in their scope. These field experiments evaluate the effect of a single content factor while controlling for others. Further, these are conducted on a small number of advertisements.%
Therefore, to model LTM over advertisements, we conduct the first large-scale human study on long-term advertisement memorability\footnote{We obtained the Institutional Review Board Approval to conduct the study from our institute.}. We call it LAMBDA (Long-term Ad MemoraBility DAtaset). Over two years, we conducted an LTM study involving 1749 participants across four sessions across two institutes to collect LAMBDA. We collect memorability scores over 2205 ads from 276 brands, covering 113 industries. On day 1, participants saw ads, and after a lag time of at least one day, they answered questions testing their brand recall, ad recall and recognition, scene recall and recognition, and audio recall (\S\ref{sec:Annotation Protocol}). Next, we average the brand recall scores across participants and compute the average long-term ad memorability scores. Then, we use these scores to train machine learning models to predict long-term ad memorability.

\textbf{How can we model customer memory?} To model customer memory, we design a novel architecture, Henry\footnote{We name the model Henry in honor of the immense contributions by the patient Henry Molaison (H.M.) \cite{squire2009legacy}. An experimental surgery conducted on him resulted in the discovery of the distinct regions responsible for LTM and STM.} (Fig.~\ref{fig:memorability-model}), incorporating world-knowledge from large language models (Llama \cite{touvron2023llama}), visual knowledge from vision encoder (EVA-CLIP \cite{sun2023eva}) and specialized perception modules covering visual and cognitive knowledge about the ad. The world knowledge helps Henry to understand the semantics of the ad, the brand knowledge and consolidate them with the visual semantics from the ad. The visual encoder helps the model to ``see'' the ad. We convert the visual encoder embeddings to language space using QFormer \cite{li2023blip} and further augment them with specialized ``verbalizations'' involving visual scene descriptors like visual caption, optical character recognition (OCR), automatic speech recognition (ASR), and cognitive descriptors like emotion and scene complexity scores, which help the model ground the visual and cognitive knowledge in the LLM's world knowledge. We train the model on our LTM data samples and obtain higher than human consistency scores. Further, we train Henry on other short and long term image and video memorability datasets in the literature - LaMem, MemCat, SUN, Memento10k, MediaEval, and obtain state-of-the-art performance on all of them. We also show that Henry performs well on unseen datasets in zero-shot settings, performing better than models specifically trained on those datasets.

\textbf{How to generate memorable Ads?}\label{ref:Generation Task} One of the primary goals of modeling content memorability is to generate more memorable content. The task of generating more memorable ads is given the ad description containing the brand and campaign title to generate the ad scenes and dialogues. However, there is no data in the literature for this task. Therefore, we turn to synthetic data generation and LLM-as-a-judge paradigm \cite{khandelwal2023large,zheng2023judging}.
We first collect a large-scale advertisements dataset, collecting brand name, ad text, time, ad content, and channel. Then, we use Henry as a judge to simulate memorability on the collected ads. We ultimately get a dataset of 5 million advertisements with their automatic speech transcripts, OCR, automatically detected objects, colors, aesthetics, captions, emotions, logos, and memorability scores. We call this dataset UltraLAMBDA. We then select high memorability samples from UltraLAMBDA to train Llama-13B to generate memorable ads. Finetuning LLama for two iterations on this automatically constructed dataset yields an improvement of 44\% in memorable ad generation.

Our main contributions are summarized as follows:
\begin{itemize}[leftmargin=*]
    \item We release the first large-scale dataset, LAMBDA, on long-term advertisement memorability involving more than 1700 participants. We collect memorability scores over 2205 ads from 276 brands (157/276 brands are from SnP 500), covering 113 industries. Further, we introduce a new protocol to measure customer memory of brands (\S\ref{sec:Annotation Protocol}).
    \item We design a novel model, Henry, which can model both STM and LTM and can incorporate scene understanding, brand knowledge, and speech (\S\ref{sec:Predicting Ad Memorability}). Henry achieves state-of-the-art performance on eight literature image and video memorability datasets (\S\ref{sec:results and discussion}). Further, we show that Henry performs well on unseen datasets in zero-shot settings.
    \item We propose the task of memorable ad generation. We release the first large scale ad dataset, UltraLAMBDA, consisting of 5 million ads with their automatically extracted content labels like ASR, captions, OCR, emotions, and memorability scores assigned by Henry. Using UltraLAMBDA, we first show that large LLMs like GPT-3.5 and 4 are unable to generate memorable content. Then, we train Henry to progressively generate more memorable ads resulting an average improvement of 44\% in memorability scores (\S\ref{sec:Generating Memorable Ads}). Through this, for the first time in literature, we also show the use of synthetic data on a task for which no large scale data exists.
    \item We conduct an extensive set of experiments on memorability prediction, showing the effects of LTM on STM modeling and vice-versa, and the effects of changing world-knowledge with time, scene understanding, brand knowledge, and speech on memorability modeling (\S\ref{sec:results and discussion}).\\
\end{itemize}

\begin{table*}[h]
\begin{center}\small
\resizebox{\textwidth}{!}{%
 \begin{tabular}{cp{1.8cm}p{5.5cm}p{1.5cm}p{5.5cm}}
     \toprule
     \small
      Image & Semantic Category & Verbalization & Semantic Category & Verbalization\\ 
    \cmidrule(r){1-1}\cmidrule(lr){2-2}\cmidrule(l){3-3}\cmidrule(l){4-4}\cmidrule(l){5-5}
     \multirow{10}{*}{\raisebox{-\totalheight}{\includegraphics[width=0.25\columnwidth]{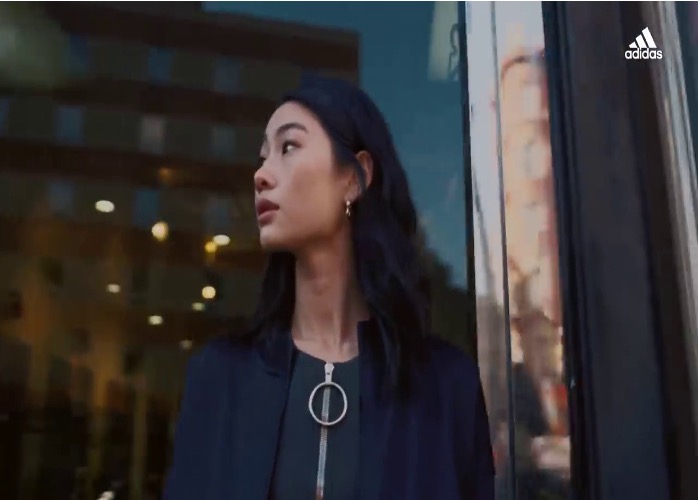}}} & \textbf{OCR} & The text shown in the scene is ``\textcolor{red}{Adidas}''. & \textbf{Clutter} & The clutter in the scene is \textcolor{red}{low}.\\
    & \textbf{ASR} & The audio in the scene is ``\textcolor{red}{To take hold of the world’s spotlight overnight}''.& \textbf{Photo Style} & The photography style of the scene is \textcolor{red}{commercial photography}.\\
    & \textbf{Human Presence} & The scene has \textcolor{red}{1 person with prominent face}.& \textbf{Emotion} & The emotion of the scene is \textcolor{red}{ambitious, determined}.\\

    & \textbf{Caption} & The scene shows \textcolor{red}{a young woman sitting in a glass door looking out}.    & \textbf{Aesthetics} & The image has \textcolor{red}{medium} aesthetic value.\\
    & \textbf{Colors} & The foreground colors of the scene are \textcolor{red}{Black, Dark Brown, Dark Blue, Dark Gray, Mud Green} and the background colors are \textcolor{red}{Dark Blue, Black, Dark Brown}. The dominant tone of the scene is \textcolor{red}{neutral}.
    & \textbf{Object Tags} & This scene is categorized by the tags: \textcolor{red}{person, woman,  blazer, facing, template, fashion, street fashion, cold, client, cardigan, sweat}.\\
    & \textbf{Audio Type} & The scene has \textcolor{red}{music and speech}. & \textbf{Logo Presence} & There is \textcolor{red}{a logo} in the scene.\\\hline
  \end{tabular}}
  \caption{To augment the scene understanding of LLM, we verbalize video scenes and images using a diverse set of cognitive and perception tools and pass it to the LLM in the format shown in the table. For image memorability datasets, we use the following semantic categories: caption, color, photo style, emotion, clutter, human presence, object tags, OCR, and aesthetics. For video scene memorability datasets, we use the following semantic categories: caption, color, emotion, human presence, object tags, ASR, OCR, Audio-type, Logo-presence. We use the following models to extract the features: OCR \protect\cite{du2020pp}, clutter \protect\cite{khurana2023synthesizing}, ASR \protect\cite{radford2022robust}, Photo style \protect\cite{li2023blip}, human presence \protect\cite{liu2023grounding}, emotion \protect\cite{singh2023video}, caption \protect\cite{li2023blip}, aesthetics \protect\cite{ke2023vila}, colors \protect\cite{Qin_2020_PR}, object tags \protect\cite{zhang2023recognize}, audio-type \protect\cite{giannakopoulos2015pyaudioanalysis}, and logo presence \protect\cite{zhang2023recognize}. Black colored text is the verbalization template, and \textcolor{red}{red} text indicates the model outputs.
  \label{table:scene-verbalization-format}}
\end{center}
\end{table*}

\section{LAMBDA Protocol, Study \& Insights}
\label{sec:Long-Term Advertisement Memorability Dataset}
We first give an overview of LAMBDA data collection process and the annotation protocol. We also present some interesting characteristics LAMBDA exhibits about LTM.

\subsection{Video Collection}
\label{sec:Video Collection}
In contrast to previous video memorability works where videos were soundless and only of action videos \cite{newman2020multimodal, cohendet2019videomem}, the videos in our dataset come from multimodal ads released on YouTube channels of 276 major brands covering 113 industries%
. We collect 2205 such ads spanning over the years 2008-2023. The videos have an average duration of 33 seconds. Out of all the videos, 2175 have audio in them. The collected advertisement videos have a variety of characteristics, including different scene velocities, human presence and animations, visual and audio branding, a variety of emotions, scene complexity, and audio types.

\subsection{Annotation Protocol}
\label{sec:Annotation Protocol}
At the outset, participants are given a preliminary questionnaire aimed at establishing their brand-related interactions and media consumption habits. Participants are given a list of fifteen randomly chosen brand options and are asked to choose those they recall encountering advertisements for during the current year. Subsequently, participants are presented with another set of fifteen brands and are instructed to identify those for which they have personally utilized products within the same timeframe. 

In addition, participants are asked about their utilization of ad-blocking software and their Youtube subscription. The questionnaire further captures participants' digital media habits, including the division of their time spent on YouTube between mobile and web platforms and their preferred channels for acquiring information about new products and brands.

Following the questionnaire, participants proceed to the core segment of the study, where they are shown 11 advertisements sequentially. Notably, the eleventh advertisement is deliberately repeated for half of the participants, while it is unique for the other half. To ensure participant engagement, attention-check questions are placed between every two to three advertisements. These questions are common sense questions like ``How many legs does a cow have?''. If the participant fails to answer the question within 10 secs, they are requested to rewatch the video. After the 11th video, participants are asked if they recollect watching the ad in the span of the study. Interestingly, 15\% participants were not able to recognize the repeated video correctly.

\begin{table*}[!h]
\vspace*{-0.2in}
\centering
\resizebox{0.99\textwidth}{!}
{%
\centering \scriptsize
\begin{tabularx}{1.0\textwidth}{>{\hsize=0.9\hsize}X|>{\hsize=0.3\hsize}X>{\hsize=0.3\hsize}X>{\hsize=0.3\hsize}X>{\hsize=0.3\hsize}X|>{\hsize=0.35\hsize}X>{\hsize=0.3\hsize}X>{\hsize=0.3\hsize}X>{\hsize=0.35\hsize}X}
\hline
\multirow{2}{*}{\textbf{Models}} & \multicolumn{4}{c|}{\textbf{Image Datasets}} & \multicolumn{4}{c}{\textbf{Video Datasets}} \\
& \textbf{Lamem} & \textbf{Memcat} & \textbf{SUN} & \textbf{Merged} & \textbf{Memento10k} & \textbf{VideoMem} & \textbf{MediaEval} & \textbf{\makecell{LAMBDA}} \\
\hline
{Human Consistency} & 0.68 & 0.78 & 0.75 & - & 0.73 & 0.61 & - & 0.55 \\
{10-shot GPT-3.5} & 0.29 & 0.18 & 0.15 & - & 0.07 & 0.06 & 0.06 & 0.06 \\
{Regression 
using ViT feats (ViTMem)} & 0.71 & 0.65 & 0.63 & \valgood{0.77} & 0.56 & 0.51 & - & 0.08 \\
{Current Literature SOTA} & 0.71 & 0.65 & 0.68 & \valgood{0.77} & 0.67 & 0.56 & 0.46 & - \\
{Henry trained on individual datasets} & \valbest{0.74} & \valbest{0.82}  & \valgood{0.73}  &-  & \valbest{0.75}  & \valbest{0.64}  & \valbest{0.50}  & \valbest{0.55}  \\
{Henry trained on all (combined) datasets} & \valgood{0.72} & \valgood{0.79} & \valbest{0.76} & \valbest{0.79} & \valgood{0.72} & \valgood{0.60} & \valgood{0.48} & \valgood{0.52} \\
\hline
\end{tabularx}%
}
\caption{Results of Henry (our model) on eight datasets compared with the current best models reported in the literature and GPT-3.5. Human consistency values are also listed in the top row for reference. It can be observed that our model achieves state-of-the-art performance across all datasets. Best models are denoted in \valbest{green} and runner-ups in \valgood{blue}.
References for the seven literature SOTA models in the format \{\texttt{dataset: SOTA model citation}\} are: LaMem: \cite{hagen2023image}, MemCat: \cite{hagen2023image}, SUN: \cite{fajtl2018amnet}, Merged Image datasets: \cite{hagen2023image}, Memento10k: \cite{Dumont_2023_CVPR}, VideoMem: \cite{Dumont_2023_CVPR}, MediaEval: \cite{DBLP:conf/mediaeval/LuW21}
}
\label{table:memorability-main-results}
\end{table*}

The memorability test included 1,749 participants: 971 in a take-home setting and 778 in an auditorium. Take-home participants received questionnaires via email 24 hours after exposure, with responses accepted within a 72-hour window. Auditorium participants completed questionnaires at intervals of 24, 36, or 72 hours, with equal distribution across these time points. The questionnaire assessed two types of memory: brand recognition and ad recall. For recognition, participants identified previously encountered brands from a randomized list of 20. For recall, they described the advertisements associated with the brands they recognized in the earlier prompt\footnote{The complete questionnaire for participant one is given in \S\ref{sec:Memorability Questionnaire}.}.

The average memorability score was 67.5\% (SD=13.6\%). To assess human consistency, we divided the participant pool into two independent halves and measured the agreement between memorability scores from one half and the other. Across 25 random split-half trials, we obtained a Spearman’s rank correlation ($\rho$) of 0.77 for brand recall (compared to $\rho=0.68$ for images \cite{khosla2015understanding}, $\rho=0.616$ for videos in \cite{cohendet2019videomem}, and $\rho=0.73$ in \cite{newman2020multimodal}). The estimated sensitivity index ($d'$) for participants was calculated as 1.848 \cite{Wikipedia_SensitivityIndex}.

\subsection{What makes an Ad memorable?}
\label{sec:What makes an Ad memorable?}
Among the many reasons why an ad might be memorable, we investigate the following factors: \textbf{brand factors} (\textit{viz.,} brand popularity, industry), \textbf{content factors} (\textit{viz.,} video emotion, scene velocity, length, speech to silence ratio), \textbf{customer-content interaction factors} (\textit{viz.,} time of seeing the video, order in which the video was seen, time difference between watching the video and recalling the brand), and \textbf{customer behavior factors} (\textit{viz.,} average relevance of the brand and video popularity).

\textbf{Content Factors}: %
Previous studies like \cite{isola2011makes,newman2020multimodal} have investigated the effect of pixel statistics like color and hue, saturation, and value, scene semantics like the number of objects, the area occupied by objects on memorability. In general, low-level semantic features have no correlation with memorability, but higher-level features like the type of scenery has some correlation. For instance, Newman \textit{et al.} \cite{newman2020multimodal} found that videos with
people, faces, hands, man-made spaces, and moving objects are, in general, more memorable than those with outdoor landscapes or dark and cluttered content. Since only our dataset has videos with cognitive features like emotions and are also non-silent, we extend the previous analysis to find the effect of speech and emotion on memory. Fig.~\ref{subfig:speech vs recall} shows the effect of speech. We observe that percentage of speech in a video, presence of music, and type of music have a very little correlation with long term memory. On the other hand, emotions primarily depicted through speech in ads can explain memorability. We see in Fig.~\ref{subfig:emotion vs recall} that negative emotions are more memorable than positive emotions. Further, %
we find that video length has little effect on memorability (Fig.~\ref{subfig:length vs mem}), but scene velocity has a slightly positive correlation with memory (Fig.~\ref{subfig:velocity vs recall}).

\textbf{Interaction Factors:} Memorability may also depend on the time of the day the ad was seen. However, we find that the time of day of watching has almost no effect on the memorability of the ad (Fig.~\ref{subfig:avg recall by time}). It may be expected that memorability decays as time passes. We plot the forgetting curve for ads in Fig.~\ref{subfig:delay vs recall} measuring brand recognition against time elapsed between video viewing and memory assessment. The forgetting coefficient of ads is 0.18, notably than action videos \cite{cohendet2019videomem}. The difference likely arises due to differences in protocols. Cohendet \textit{et al.} (2019) \cite{cohendet2019videomem} used a two-stage memory protocol in which participants did both short-term and long-term recall, thus enhancing their long-term recall. 
Next, we investigate the effect of the order in which the video was watched with its memorability (Fig.~\ref{subfig:avg recall by position}). We see that order of videos seen has little impact on video memorability, with a slight bias in favor of the initial and last ads. %

\textbf{Customer Behavior Factors}: It might be possible that the videos which are liked more are remembered more. To investigate this, we test the correlation of popularity as measured by the ratio of Youtube video likes to views with memorability. We see that there is a positive correlation between video popularity and memorability (Fig.~\ref{subfig:popularity vs recall}). Further, in the study, we asked the participants to select the brands they have personally used from a set of 15 randomly chosen brands and similarly choose brands they have seen ads for. To prevent any systematic bias, the brands asked in this question are independent of the brands shown the next day. We plot thus collected brand relevance values with brand recall in Fig.~\ref{subfig:relevance vs recall}. We see that average brand relevance is strongly correlated with average recall (coeff= 0.53), where entertainment, corporate, and food and beverage sectors, which are quite popular brands in a student population are the most remembered, while the others are less remembered (Fig.~\ref{subfig:recall by sector}).

\section{Predicting Ad Memorability}
\label{sec:Predicting Ad Memorability}

In this section, we focus on predicting memorability - both long-term and short-term for both videos and images. We pose memorability prediction as a problem that needs (a)~\textit{visual knowledge} to identify and understand visual concepts across images and videos like shapes, colors, objects, and scenes, (b)~\textit{cognitive knowledge} relevant to marketing, for example, ad emotions, scene complexity, scene aesthetics, and (c)~\textit{world knowledge} to relate the captured visual and marketing concepts to real-world concepts capturing their function, use, and interaction patterns. %
For instance, when Airbnb\footnote{See Fig.~\ref{fig:airbnb-ad}) for the ad} shows an adult female and a male with the text, ``Our guest room is paying for our wedding''; it denotes a couple saying that renting out their space on Airbnb helps them sponsor their wedding \cite{kumar2023persuasion}. World knowledge captured in LLMs, together with the visual knowledge of ViT and marketing knowledge through specialized cognitive models, helps to (i)~identify the two adults as a couple, (ii)~AirBnb as a housing company, (iii)~recognize the warm emotional tone of the text, and make sense of all three concepts together. Fig.~\ref{fig:memorability-model} shows the proposed architecture of Henry. %

\begin{table*}[!t]
\centering
\resizebox{1.0\textwidth}{!}{%
\begin{tabular}{l|ll|llllllll}
\hline
\textbf{Generalization Type} &  \textbf{Train on} & \textbf{Zero-shot Testing} & \textbf{Lamem} & \textbf{Memcat} & \textbf{SUN} & \textbf{VideoMem} & \textbf{Memento10k} & \textbf{LAMBDA} \\ \hline
Memory-type & Short-term & Long-term & - &-        &-     & 0.31         & - & 0.18          \\
Memory-type & Long-term & Short-term &0.06       &0.08        &0.07     &0.15          &0.1         &-           \\
Modality & Videos  & Images &0.55     &0.65        &0.55     &-          &-         &-           \\
Modality & Images & Videos              &-       &-        &-     &0.44          &0.54         &0.09           \\
Brands & All except 20 brands & Left-out 20 brands &-       &-        &-     &-          &-           &0.42           \\
Dataset & All except Memento & Memento &-      &-        &-     &-         & 0.59           &-           \\
Dataset & All except Memcat & Memcat   &-       &0.68        &-     & -         &-         & -          \\\hline
\end{tabular}%
}
\caption{Ablation across data to understand how memorability prediction generalizes across the type of memory, datasets, modality (image/video), and brands. The reported values are correlations between model and human memorability scores. A few trends can be observed from the table: (i)~STM generalizes better on LTM in zero-shot than vice versa (rows 1 and 2), (ii)~Henry trained on either videos or images generalizes to both (rows 3 and 4), (iii)~There is a significant performance loss in modeling memorability for brands not seen during training (row 5), (iv)~Zero-shot generalization to Memento (video) and Memcat (image) is near to the current trained state of the art literature models on Memento \cite{Dumont_2023_CVPR} and Memcat \cite{hagen2023image} (rows 6 and 7).}
\label{table:data-ablation}
\end{table*}

\subsection{Encoding Multimodal Content}
\label{sec:Encoding Visual Content}
The primary goal of this step is to effectively leverage the ``world-knowledge'' capabilities of the pre-trained LLM. We choose Llama \cite{touvron2023llama} as our base LLM. We employ two techniques to convert visual data into language: encoding visual frames into the LLM space and verbalizing cognitive concepts into language space. We detail the two steps next.

\textbf{Sampling Frames:} %
We detect scene changes by analyzing changes in HSV intensity and edges in the scene, with a 0.3 threshold. We choose the threshold value from the 30-degree rule inspired by the concept of jump-cut avoidance in cinematography \cite{arev2014automatic,friedman2004knowledge}. The 30-degree rule can be formulated as follows: after a ``cut'' (camera stops and re-starts shooting), the camera angle must change by at least 30 degrees.
For dominant frame selection common blur/sharpness heuristics fail in presence of text in image. So we extract the frame with the least changes using \cite{xu2022gmflow}.

\textbf{Encoding Into Language Embedding Space:} To give visual knowledge to Henry, we use EVA-CLIP visual embedder \cite{sun2023eva}. We find that Global Multi-Head Relation Aggregator (GMHRA) \cite{li2021uniformer} helps aggregate the ViT's information better across the time dimension. Next, to effectively leverage the LLM's rich language representations, we use a pretrained Q-Former from BLIP-2 \cite{li2023blip} with an extra linear layer and additional query tokens to convert from visual tokens to language tokens.

\textbf{Verbalizing Cognitive, Experimental, Visual Concepts}
\label{sec:Verbalizing Visual Content}
While visual content encodings are a good representation of the visual characteristics of the image, we find that they are still unable to capture rich cognitive and semantic information present in images. Therefore, to augment the cognitive understanding of the LLM, we verbalize the frame semantic information using the set of features that came out important in our memorability analysis (Fig.~\ref{fig:correlation-graphs}) \cite{bhattacharya2023video,singh2024llava}. The cognitive and visual features are given in Table~\ref{table:scene-verbalization-format} and Listing~\ref{listing:memorability-prediction-verbalization-format}. We find that our cognitive verbalization helps ground the visual perception of LLM in the marketing concepts of the image, helping in downstream prediction performance (Table~\ref{table:architecture-ablation}). %

\subsection{Two-Stage Training}
\label{sec:Two-Stage Training}
We do two-stage training where in the first stage, we utilize the Webvid \cite{bain2021frozen}, COCO caption \cite{chen2015microsoft}, Visual Genome \cite{krishna2017visual}, CC3M \cite{sharma2018conceptual}, and CC12M \cite{changpinyo2021conceptual} datasets to align the visual encoder embeddings with LLM via a large-scale pretraining approach. In the second stage, we train the model with high-quality memorability instructions prepared by following the approach described in the last paragraphs. Henry takes the concatenated inputs, representing the contextual information, and is trained to predict the memorability score of the given image or video within the range of 00 to 99 (see Listing~\ref{listing:memorability-prediction-verbalization-format}). The memorability score of a video, is the percentage of times the participants recall the video correctly (we normalise it to an integer value between 00 and 99 to facilitate the LLM training). During training, the LLM predicts from the complete vocabulary, while during inference, we use the softmax function over numeric tokens only to obtain a number.

\subsection{Results and Discussion}
\label{sec:results and discussion}
We conduct extensive experiments on all literature datasets, covering both videos and images, STM and LTM. We compare Henry\footnote{Computing infrastructure used to conduct the experiments along with hyperparameters are given in \S\ref{sec:experimental-details}. All experiments are conducted with three random seeds and averages are reported.} with the current state-of-the-art models in the literature across eight datasets, including 10-shot GPT-3.5 (text-davinci-003) \cite{ouyang2022training} where we provide GPT with the same verbalization (for 10 examples), as we provided to Henry, as well as with prior regression based methods using features extracted from ViT L-14 \cite{hagen2023image}. Results are shown in Table~\ref{table:memorability-main-results}, which demonstrate that Henry outperforms all the seven models in the literature across all the seven datasets. We also conduct extensive ablations to understand the effect of different kinds of data and architectural choices. %

\begin{table*}[!t]
\centering
\resizebox{0.99\textwidth}{!}{%
\begin{tabular}{lllllll}
\hline

& \textbf{Lamem} & \textbf{Memcat} & \textbf{VideoMem(ST)} & \textbf{Memento10k}  & \textbf{VideoMem(LT)}  & \textbf{LAMBDA} 
\\ \hline
\makecell{Henry on individual datasets} & 0.74 & 0.82   &0.64 & 0.75  & 0.48   & 0.55  \\
Henry vision only &0.20       & 0.17       &0.17       &0.21        &0.15         &0.11  \\
Henry language only &0.51       & 0.53       &0.42       &0.54        &0.37         &0.44  \\
Henry -object tags &0.67       &0.71        &0.57       &0.69       &0.46         &0.52         \\
Henry -colors &0.65       &0.74        &0.55      &0.67       &0.45         &0.51         \\
Henry -emotion &0.71       &0.78        &0.61      & 0.73       &0.42         & 0.46         \\
Henry -aesthetics &0.72       &0.79        &0.61       &0.71       &0.46         &0.53         \\
Henry -clutter &0.73       &0.81        &0.60       &0.74       &0.45         &0.53     \\
Henry -asr &-       &-        &-       & -       &-         & 0.46         \\
Henry -asr-emotion &-       &-        &-       & -       &-         &0.42         \\
Henry on Silent Ads&-       &-        &-       & -       &-         &0.56        \\
Henry on Ads with audio&-       &-        &-       & -       &-         &0.52        \\

\hline%
\end{tabular}%
}
\caption{Ablation across architectural choices. ``-'' denotes non-speech dataset. A few trends are visible from the table: (i)~Despite having a vision branch, object tags and colors have a net positive impact on the overall performance (rows 2,3,4), (ii)~For LTM (LAMBDA, VideoMem (LT)), dropping cognitive features such as emotion, aesthetics, and clutter cause a larger performance drop than dropping visual features such as objects and colors. The trend is the opposite for STM (Lamem, Memcat, VideoMem (ST), Memento10k).}
\label{table:architecture-ablation}
\end{table*}

\subsection{Ablation of Data and Architectural Choices on Memorability Prediction}
\label{sec:Ablation of Data and Architectural Choices on Memorability Prediction}
We also conduct extensive ablations to understand the effect of different kinds of data and architectural choices. Tables~\ref{table:memorability-main-results} and Table~\ref{table:data-ablation} show the data ablations. We see that combining datasets actually worsens the performance across all the datasets except the SUN dataset. Further, we find that in zero-shot settings, STM helps in predicting LTM relatively much better than vice versa. This corroborates with the studies in psychology which show that for a content to get committed to LTM, it has to pass through STM \cite{norris2017short}. Therefore, content memorable, according to STM, has an effect on LTM but, interestingly, not vice versa. %
Further, we observe that Henry loses performance for unseen brands. This underscores the importance of scaling the study across more brands. Next, we evaluate the impact of various architectural choices  (Table~\ref{table:architecture-ablation}). We find that Henry's vision branch is not strong enough by itself to produce good results. Cognitive features that were found important in our study also improve prediction performance. Low-level features like objects and colors have the maximum impact on STM, but higher-level features like emotion, ASR, and aesthetics have a higher impact on LTM.

\section{Generating Memorable Ads}
\label{sec:Generating Memorable Ads}

\begin{figure}[!ht]
    \centering
    \includegraphics[width=0.48\textwidth]{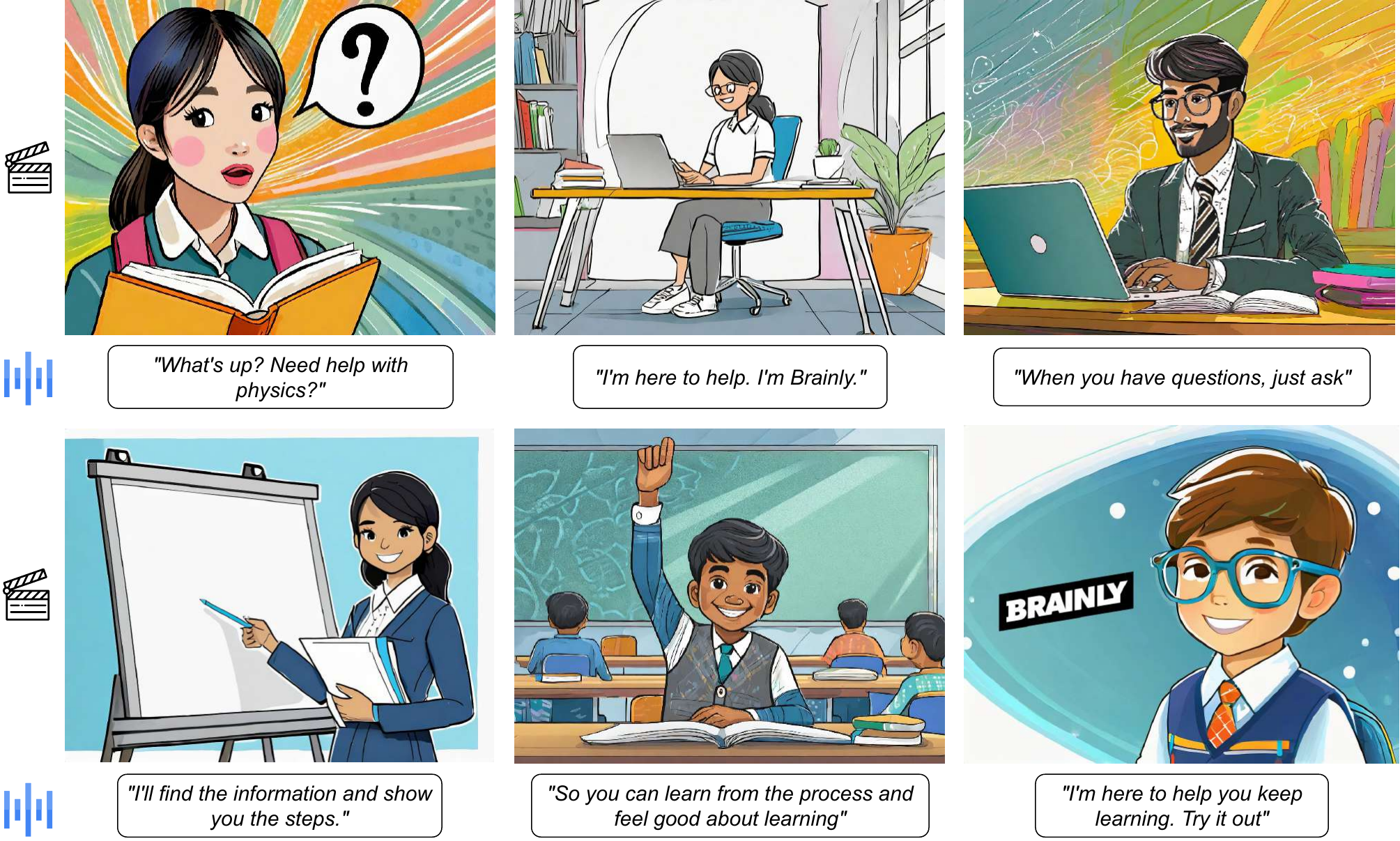}
    \caption{Henry-SEED Prompt: \textit{Generate the detailed description of a 30-second memorable advertisement titled "Brainly Keep Learning 30sec Final 16x9" for the brand Brainly}. Link to the original ad: \url{https://www.youtube.com/watch?v=kytRXyWXivU} Original Memorability score: 85. Memorability score of Generated Ad: 99.}
    \label{fig:brainly-generated-ad}
\vspace{-3mm}
\end{figure}

\begin{figure}[!t]
    \centering
    \includegraphics[width=0.98\columnwidth]{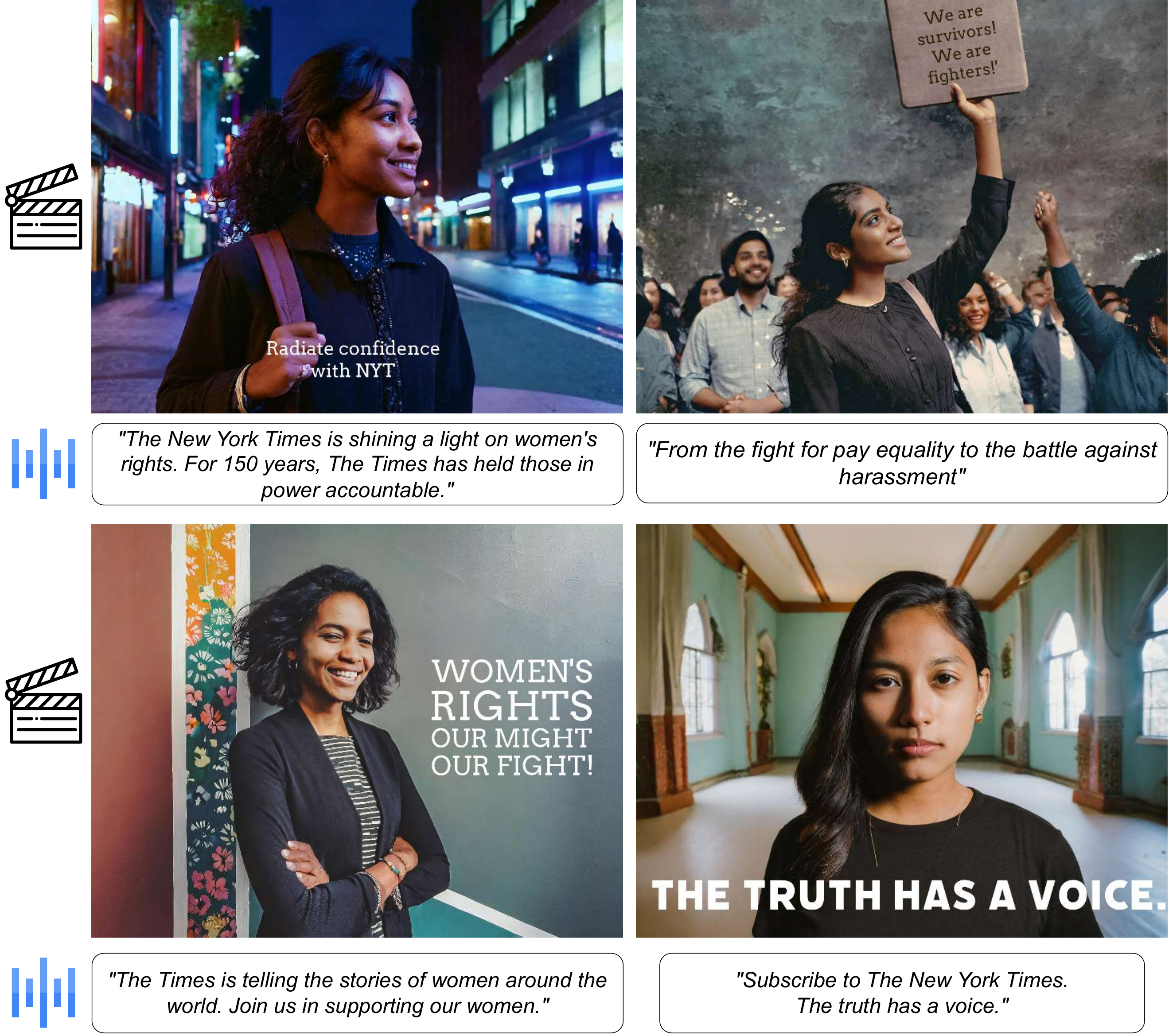}
    \caption{Henry-SEED Prompt: \textit{Generate the detailed description of a 50 second memorable advertisement titled "Shining a Light on Women’s Rights | The Truth Has a Voice | The New York Times" for the brand The New York Times} Link to the original ad: \url{https://www.youtube.com/watch?v=bPblzhUzTeg} Original memorability score: 65. Memorability score of Generated Ad: 91.}
    \label{fig:adgen-NYT}
\end{figure}

\begin{figure*}[!ht]
    \centering
    \includegraphics[width=0.99\textwidth]{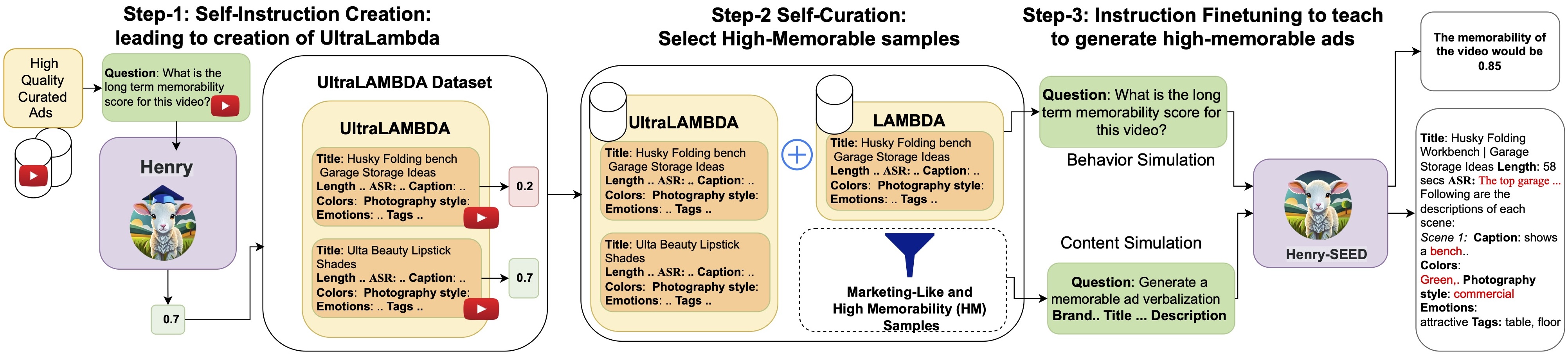}
    \caption{Overview of our SEED method for memorable ad generation. Our self-alignment consists of three steps: (i)~\textbf{Self-instruction creation}: We first collect 5 million high-quality ads from YouTube, Facebook, and other mediums. Henry (trained on the complete train+test sets of LAMBDA) is then used to rate this curated set in an LLM-as-a-Judge fashion. (ii)~\textbf{Self-curation}: We select marketing-like and high-memorability samples from the UltraLAMBDA and LAMBDA datasets. (iii)~Instruction fine-tuning: Henry-SEED is trained on the self-curated set using two tasks: Behavior Simulation and Content Simulation. 
    \label{fig:Henry-SEED}}    
\end{figure*}

\begin{figure}[!b]
    \centering
    \includegraphics[width=0.95\columnwidth]{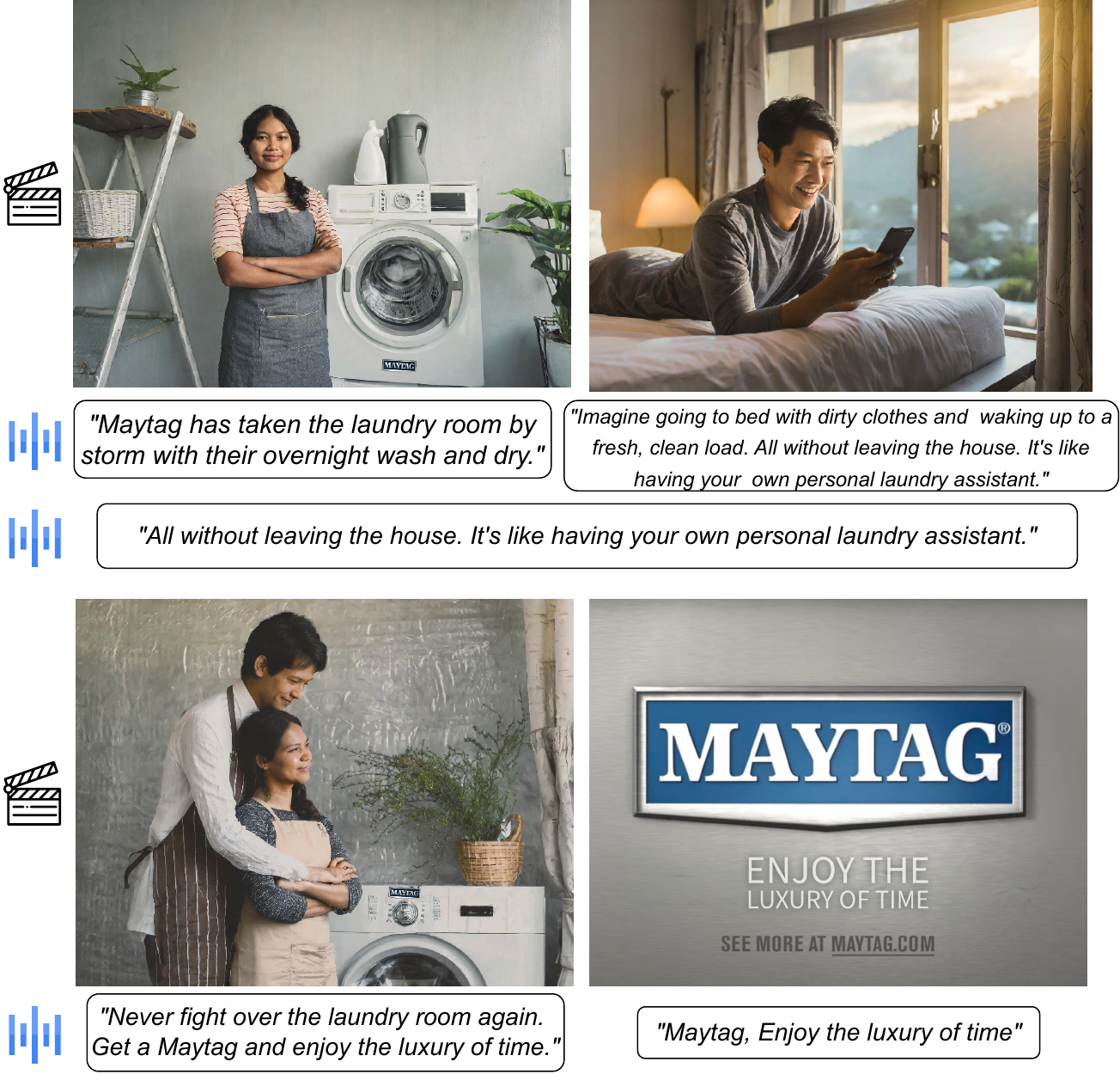}
    \caption{Henry-SEED Prompt: \textit{Generate the detailed description of a 18 second memorable advertisement titled "Maytag Overnight Wash and Dry" for the brand Costco}. Link to the original ad: \url{https://www.youtube.com/watch?v=uT72lJhUUS0} Original memorability score: 76. Memorability score of the generated Ad: 83}
    \label{fig:maytag-generated-ad}
\end{figure}

We introduce the new task of memorable ad generation. Given inputs like a brand name, a brief campaign description, and the desired ad duration, the goal is to generate a memorable ad featuring scene descriptions, characters, and dialogues. While most memorability research focuses on assessing how memorable content is, little attention has been given to generating memorable content \cite{danescu-niculescu-mizil-etal-2012-hello,khosla2013modifying,siarohin2017make,goetschalckx2019ganalyze,kyle2022modulating}. This gap exists primarily due to the lack of a sufficiently large dataset for training models to generate memorable ads. To address this, we release a large-scale dataset of raw ads and propose the Self-rEwarding mEmorability moDeling (SEED) method, which leverages raw ads to create memorable ones.

\textbf{SEED method} (Fig.~\ref{fig:Henry-SEED}):  \textit{Step 1: Self-Instruction Creation:} We gather a dataset of 5 million raw ads sourced from social media platforms, including Facebook, Twitter, Snapchat, and YouTube. For each ad, we collect the brand name, ad title, links, captions, dates, and ad assets (videos and images).

\textit{Step 2: Self-Curation:} Since these ads are publicly sourced, we employ few-shot Mistral-7B \cite{jiang2023mistral} to clean and filter the ads, ensuring they are marketing-focused, semantically relevant, and use proper language (Listing~\ref{listing:Mistral Prompt for Ad Filtering}). We then automatically label the ads with cognitive features critical for modeling memorability (Table~\ref{table:scene-verbalization-format}). %
Subsequently, we use Henry to label the ads for memorability scores. This results in a dataset we call \textit{UltraLAMBDA}, from which we select high-memorability ads with scores above 65.

\textit{Step 3: Instruction Fine-Tuning:} We then train LLaMA-13B to perform two tasks simultaneously: behavior simulation (predicting ad memorability based on ad content; Listing~\ref{listing:memorability-prediction-verbalization-format}) and content simulation (generating ad scenes and dialogues from a brand name, ad title, and required duration; Listing~\ref{listing:advertisement-generation-prompt-Henry-SEED}). We refer to the model trained using the SEED process as Henry-SEED (Fig.~\ref{fig:Henry-SEED}).

\begin{table*}[!t]
\vspace*{-0.2in}
\centering
\resizebox{0.99\textwidth}{!}{%
\begin{tabular}{lllllllll|lll}
\hline
\textbf{Model} & \textbf{\# Params} & \textbf{Training} & \textbf{Dataset} & \makecell{\textbf{High Quality}\\\textbf{Mem Samples}} & \multicolumn{4}{c}{\textbf{$\Delta$ Memorability}} & \multicolumn{3}{c}{\textbf{Ad-Quality}} \\
&& & & & \makecell{Low} & \makecell{Med} & \makecell{High} & Avg & \makecell{GPT-4\\Consistency} & \makecell{GPT-4\\Preference} & \makecell{Human-\\Preference}\\\hline
GPT-4 5-shot &$>$175B & ICL & $LAMBDA_{High}$ & 5 & +48 & \valgood{+18} & -13 & +17.6& \valbest{7.73}  & \valbest{91.3\%}  & \valgood{41.8\%}\\
GPT-3.5 5-shot&175B & ICL & $LAMBDA_{High}$ & 5 & +35 & +5 & -31 & +3 & 7.17 & 84.2\% & - \\
GPT-3.5 3-shot&175B & ICL & $LAMBDA_{High}$ & 3 & +34 & +6 & -32 & +2.6 & 6.98  & 83.1\% & - \\
Henry-SEED &13B& SEED & $UltraLAMBDA$ & 800k & +41 & \valgood{+18} & +1 & +20 & 7.34 & 74.7\% &   -\\
\midrule
Henry-SEED & 13B& SEED & $UltraLAMBDA$ + $LAMBDA_{High}$ & 820k & \valbest{+89} & \valbest{+31} & \valbest{+12} & \valbest{+44} & \valgood{7.44} & \valgood{85.6\%} & \valbest{60.48\%}\\
Henry-SEED & 13B&SEED & $LAMBDA_{High}$ & 650 & +78 & +13 & \valgood{+1} &  \valgood{+30.6} & 5.03 & 63.9\% & - \\
Henry-SEED & 13B&SEED & $UltraLAMBDA$ & 50k & +12 & +9 & -6 & +5 & 6.01 & 66.1\% & - \\
Henry-SEED & 13B& SEED & $UltraLAMBDA$ (w/o high-mem filtering) & 2M & +19 & +5 & -45 & -7 & 6.73  & 71.1\% & - \\\hline
\end{tabular}}
\caption{\textbf{Ad Generation}: Results of Henry-SEED compared with in-context-learning (ICL) GPT-3.5, 4 on Ad-Memorability and Ad generation quality. See \S\ref{sec:Generating Memorable Ads} for details of the metrics computed. We see that Henry-SEED generated ads are more memorable than ads generated using 15x larger GPT-3.5 and GPT-4. We test ad quality using GPT-4 as judge and then test the top-two models using human annotators. GPT-4 as a judge rates GPT-4 and Henry-SEED as the top two models. Subsequently, we ask humans to select between the original and generated ad stories. We observed that human annotators preferred Henry-SEED ads more than the original ads 3/5 times, while GPT-4 generated ads are preferred 2/5 times over the original ads. Further, we note that an increase in the amount of training data for Henry-SEED increases its performance across all metrics. Figs.~\ref{fig:brainly-generated-ad}-\ref{fig:maytag-generated-ad} and Listings~\ref{lst:maytag}-\ref{lst:publix} contain some qualitative samples generated using Henry-SEED. \label{tab:Henry-SEED-generating-memorable-ads}}
\vspace*{-3mm}
\end{table*}

\subsection{Evaluation} 
We assess the generated ads using four key metrics: (1)~memorability, as determined by Henry-Oracle\footnote{The Henry model trained on the complete (test+train sets) LAMBDA.}, (2)~memorability evaluated using perplexity of the generative models on ground-truth high/medium/low ads, (3)~ad quality as judged by GPT-4, and (4)~ad quality as evaluated by humans. Although content memorability is assessed by average human recall, it is important to note that humans cannot accurately predict how memorable content will be for others \cite{isola2013makes}. A true test of memorability for generated ads would require a memorability study akin to LAMBDA, which is costly and unscalable due to the number of models and generated ads. Therefore, we measure the memorability of generated ads using two approaches: Henry-Oracle and perplexity on ground truth memorable ads in LAMDA. 

In evaluation using \textit{Henry-Oracle}, the expectation is that the generated ad's memorability should be at par with high-memorable samples (score$>$65) and better than the low (score$<$44) and medium memorability samples (44$<$score$<$65). Perplexity on ground truth low and high memorable ads evaluates the generative model's propensity to generate more memorable content. A stronger model should have a lower perplexity on more memorable content than less memorable content (refer \S\ref{sec:Perplexity evaluation} for details on perplexity evaluation). 

Using \textit{GPT-4 as judge}, we test two ad-quality metrics: \textit{consistency} and \textit{preference}. 
Consistency assesses how coherent the generated story is—both internally (e.g., between dialogues) and in relation to the provided brand information and title (Listing~\ref{lst:ad-quality-consistency-prompt}). 
Preference measures how often GPT-4 favors the generated story over the original (Listing~\ref{lst:ad-quality-preference-prompt}). In \textit{human evaluation}, we ask human annotators to select between the generated and the original ad stories without revealing which is which (\S\ref{sec:Questionnaire to Gather Human Preferences over Generated Ads}). This evaluation is conducted with 20 non-expert annotators and 3 ad industry experts with over 5 years of experience in the creative industry. The expectation is that the quality of synthetic ads should be comparable to that of the original ads.

\begin{table}[!b]
\resizebox{0.48\textwidth}{!}{%
\begin{tabular}{cc|ccc}
    \hline
    \textbf{Model} & \textbf{Training} & \textbf{Low}($\uparrow$) & \textbf{Medium} & \textbf{High}($\downarrow$) \\ \toprule
    LLaVA & 0-shot & 5.08 & 5.11 & 5.39 \\ \hline
    Henry-SEED & LAMBDA$_{HIGH}$ & 6.07 & 3.01 & 2.17 \\ \hline
    Henry-SEED & UltraLAMBDA & 7.09 & 4.51 & 2.35\\ \bottomrule
    \end{tabular}}
    \caption{\textbf{Ad Generation}: Perplexity comparison (refer \S\ref{sec:Perplexity evaluation}) of LLaVA and Henry-SEED on low/medium/high memorable ads from LAMBDA test set. We see that untrained LLaVA does not favor memorable ads. Further, we note that when synthetic data is included during training, the ratio of perplexity on low and high ads grows from 2.79 to 3.01.}
    \label{table:perplexity-scores-ultralamda}
\end{table}

\subsection{Results}  
We compare the following models to generate memorable ads: LLaVA model trained on UltraLAMBDA (we refer to this model as Henry-SEED), GPT-3.5, and GPT-4. GPT-3.5 and 4 are LLMs with strong generative capabilities with high performance across many benchmarks \cite{brown2020language}.

\textbf{Evaluation of memorability of the generated ads:} Table~\ref{tab:Henry-SEED-generating-memorable-ads} compares models based on the average increase in memorability, as evaluated by the Oracle model trained on both the train and test sets. Table~\ref{table:perplexity-scores-ultralamda} presents the perplexity of LLaVA before and after training on UltraLAMBDA. Notably, Henry-SEED, trained on UltraLAMBDA, achieves significant improvements in memorability scores across all categories (Low, Medium, and High). In contrast, while GPT-4 and GPT-3.5—despite being 15x larger—enhance the memorability of ads with initially low ratings, they reduce the memorability of ads with high ratings. Table~\ref{table:perplexity-scores-ultralamda} further highlights differences between untrained and SEED-trained LLaVA. The SEED method substantially lowers perplexity on high-memorability samples. While the original LLaVA model exhibited higher perplexity for high-memorability samples, training on UltraLAMBDA reverses this trend: perplexity increases for low-memorability samples and decreases for high-memorability ones. This shift suggests that the SEED approach enhances the generation of high-memorability ads while simultaneously reducing the likelihood of producing low-memorability ones.

Importantly, UltraLAMBDA contains no overlap with LAMBDA. Neither Henry (used to label memorability for UltraLAMBDA) nor Henry-SEED (trained on UltraLAMBDA) was trained on LAMBDA’s test-set ads. Despite this, Henry-SEED demonstrates significant improvement in performance compared to GPT-3.5 and GPT-4.

\textbf{Evaluation of the quality of the generated ads:}
When comparing ad quality, we find that while GPT-4 favors its own generated ads 91.3\% of the time, Henry-SEED follows closely with an 85.6\% preference score. In human evaluations, where annotators were asked to choose between original and generated ads based on quality, Henry-SEED's ads were preferred around 60\% of the time—approximately 20\% more than GPT-4's ads.

\textbf{Qualitative Results:} Figs.~\ref{fig:brainly-generated-ad}-\ref{fig:maytag-generated-ad} and Listings~\ref{lst:maytag}-\ref{lst:publix} show some randomly sampled ad storyboards generated by Henry-SEED and Sec.~\ref{sec:Expert Feedback Collected For Generated Ads} contains some expert comments over the generated ad storyboards. These qualitative examples are generated by prompting Adobe Firefly \cite{adobefirefly} with the scene descriptions provided by Henry-SEED\footnote{Note: We do not make any changes to Henry-SEED's generation for the voice-over or the scene descriptions before passing it to Firefly.}, followed by pasting OCR from the Henry-SEED generated verbalization on top of the generated images. %
We provide visualizations for easier understanding (Figs.~\ref{fig:brainly-generated-ad}-\ref{fig:maytag-generated-ad}), along with the raw generations (Listings~\ref{lst:maytag}-\ref{lst:publix}). %

\begin{table}
\centering
\begin{tabular}{l|l|l}
\hline
Task    & LAMBDA ($\rho$) & $\Delta$ Memorability \\\hline
BS-only & 0.541 & - \\
CS-only & - & +28.41 \\
BS+CS   & 0.547 & +30.66 \\\hline
\end{tabular}
\caption{Ablation on modeling behavior simulation (BS) or memorability prediction and Content Simulation (CS) on memorable ad generation together. For memorability prediction, we again show the Spearman rank correlation on the test set similar to Table~\ref{table:memorability-main-results}; for generation, we measure the change in memorability according to Henry Oracle similar to Table~\ref{tab:Henry-SEED-generating-memorable-ads}. We observe that mixing the two tasks together increases the performance across both tasks.}
\label{table:behaviour+content simulation}
\end{table}

\begin{figure}[t]
    \centering
    \begin{subfigure}{0.485\textwidth}
        \centering
        \includegraphics[width=\textwidth]{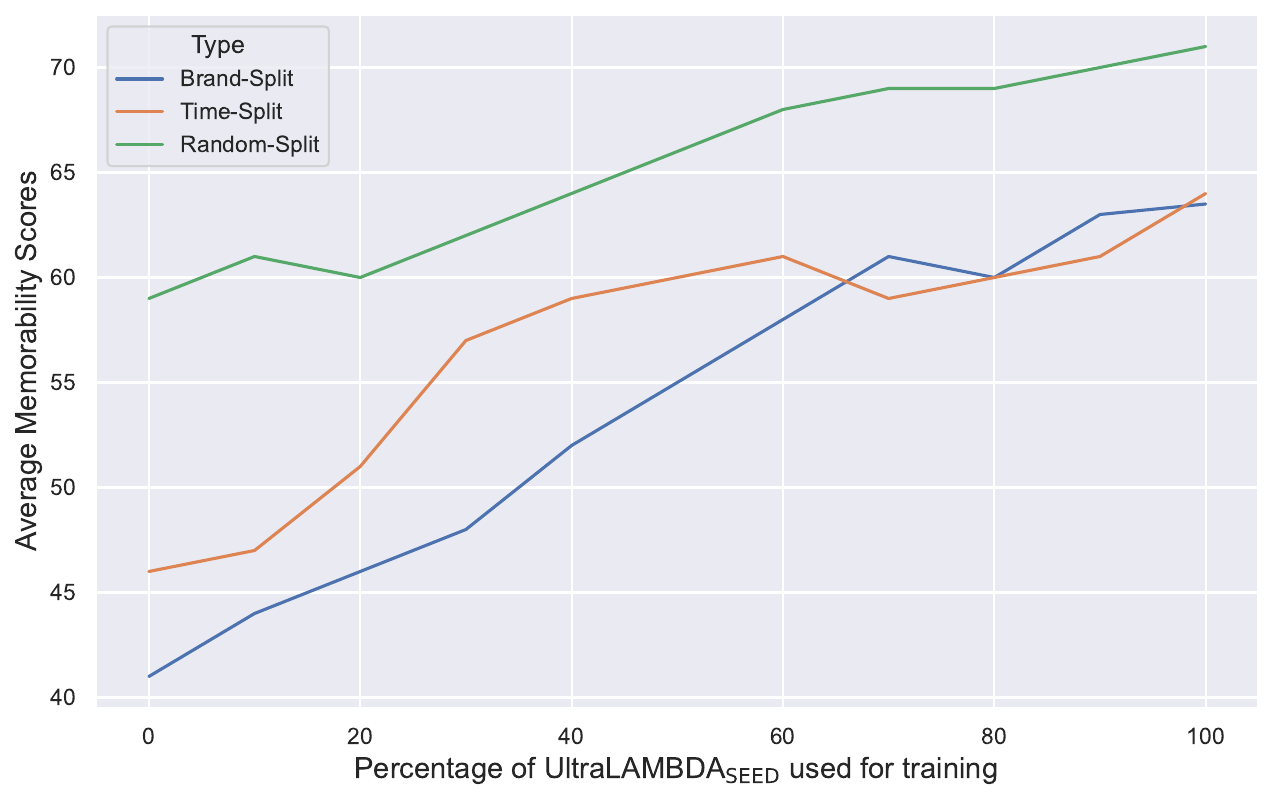}
        \label{subfig:mem vs synthetic data}
        \caption{}
    \end{subfigure}

    \begin{subfigure}{0.485\textwidth}
        \centering
        \includegraphics[width=\textwidth]{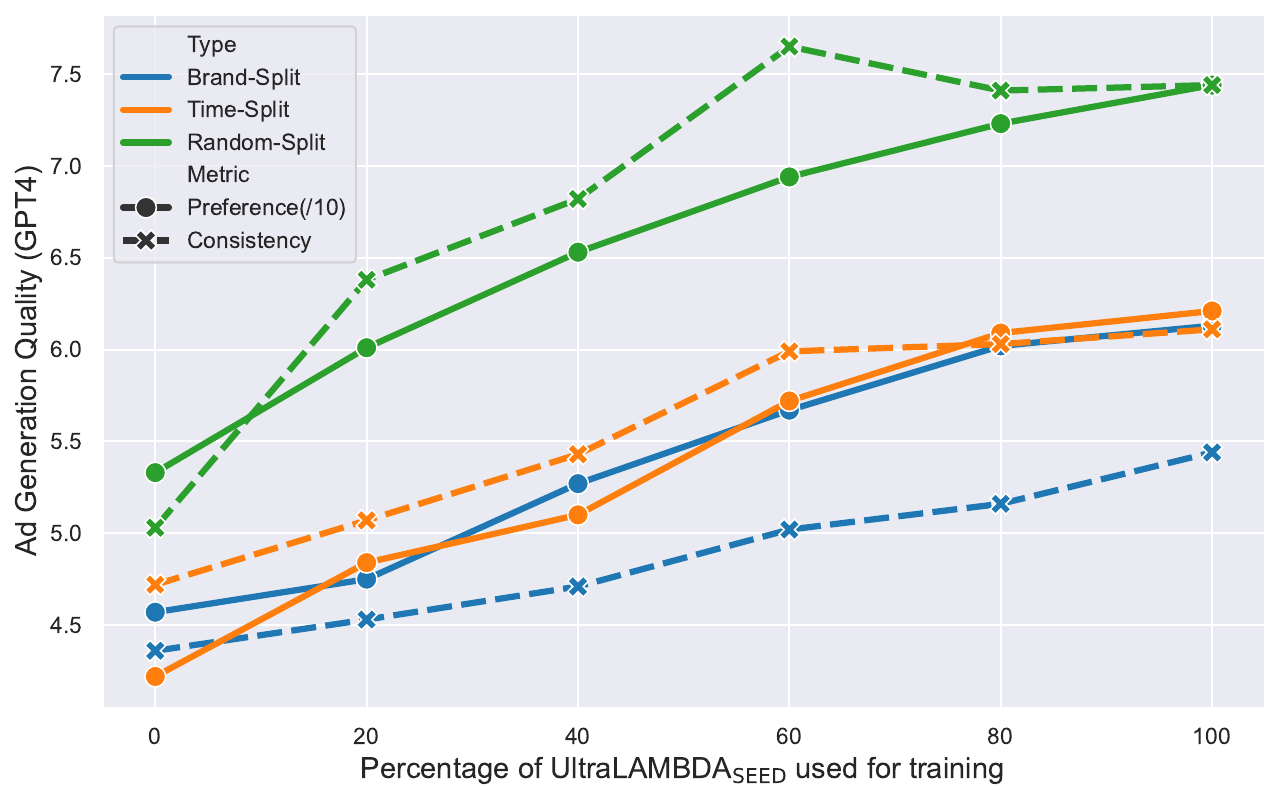}
        \label{subfig:ad-quality vs synthetic data}
        \caption{}
    \end{subfigure}
    \caption{Graphs showing the importance of the amount of synthetic data on (a)~Ad memorability score and (b)~Ad quality for the generated ads. As we can see from the graphs, both the ad memorability and quality increase with the increase in the amount of synthetic data. \label{fig:synthetic-data-vs-generation-performance}}
\end{figure}

\subsection{Ablation of Data and Architectural Choices on Memorable Ad Generation} 
\label{sec:Ablation of Data and Architectural Choices on Memorable Ad Generation}
We also run some ablation studies to find the impact of the amount of data (Fig.~\ref{fig:synthetic-data-vs-generation-performance}) and the impact of behavior simulation and content simulation tasks (Table~\ref{table:behaviour+content simulation}) on ad quality and memorability. A few trends are noticeable. Performance increases as the amount of data increases. Interestingly, the performance converges very slowly with the amount of increase in data. We test the performance in three conditions: brand-split, time-split, and random-split. In the brand-split testing, we leave some randomly chosen brands out of training and only test on them. For the time-split testing, we put a cutoff time; we train our model before that cutoff time and test on ads after that time. For the random-split testing, we test on randomly selected advertisements. Brand-split performs worse than time-split testing, indicating that brands have a higher contribution to determining memorability. This trend is observed only in ad memorability but not in ad quality. 

We also test the impact of various subsets of UltraLAMBDA on the memorability of the ads generated by Henry-SEED. Table~\ref{tab:Henry-SEED-generating-memorable-ads} shows the results. It can be seen that adding the high memorable samples from LAMBDA train set, increases the memorability of generated ads substantially. We also train LLaVA on the complete set of 2 million UltraLAMBDA ads without filtering it via Henry assigned memory labels. Interestingly, this model, while trained on 2.5 times more data than UltraLAMBDA filtered via Henry, has lesser average memorability than it.

\section{Conclusion}
\label{sec:Conclusion}
In this work, we presented the first large-scale ad memorability study and dataset, LAMBDA, measuring long-term memorability. Despite the importance that advertising plays in day-to-day, no large-scale works have tried to model long-term memorability on this multimodal content type. We then presented our model, Henry, which incorporates world and cognitive knowledge to understand the semantics of the ad content, brand, and experimental protocol, ultimately consolidating them together to predict memorability. Henry, when tested on eight datasets across the literature, spanning both short-term and long-term memorability, gets state-of-the-art performance on all of them. Next, we propose the task of generating memorable ads and release a large scale dataset UltraLAMBDA, consisting of 5 million ads for this task. We propose a new method based on self-rewarding language model to generate more memorable ads, which we call, SEED. Finetuning Henry using SEED results in an improvement of over 44\% in content memorability.

\bibliographystyle{ieee_fullname}
\bibliography{aaai23}

{\clearpage
\appendix
}

\section*{Appendix}

\section{Generation of Ads using Henry-SEED}
\label{sec:Generated Ad Samples}

Henry-SEED takes as input a prompt consisting of the title of the ad to be generated and the brand name and generates the scene-by-scene verbalization of cognitive and visual concepts and voiceover. Listings~\ref{lst:maytag}-\ref{lst:publix} gives the input and output of a few randomly chosen examples of advertisements generated by Henry-SEED are shown below. Using the output script generated by Henry-SEED, we next prompt Adobe Firefly \cite{adobefirefly} with Henry-SEED's output to generate the visualization. Next, OCR from the Henry-SEED generated verbalization is put on top of the generated images manually. We provide visualizations for easier understanding (Figs.~\ref{fig:brainly-generated-ad}-\ref{fig:maytag-generated-ad}), along with the raw generations (Listings~\ref{lst:maytag}-\ref{lst:publix}).

\begin{lstlisting}[caption={Advertisement generated by Henry-SEED for the Costco Ad. Original Ad: \url{https://www.youtube.com/watch?v=uT72lJhUUS0}},frame=single,breaklines=true,basicstyle=\tiny, label={lst:maytag}]
Input Title: "Maytag Overnight Wash and Dry" for the brand Costco


Generated Ad: "The ad is shot in landscape orientation, at a medium pace. 

Scene 1: The scene shows a person standing in front of a washer and dryer
Emotions: comforting, comfortable
Tags: appliance, laundry, laundry room, man
Voiceover: "Maytag has taken the laundry room by storm with their overnight wash and dry."
Colors: Gray, Off_White, Silver, White, Black, Dark_Gray

Scene 2: The scene shows a person lying on a bed pressing a button on a cell phone
Emotions: gift, birthday, holiday
Tags: button, catch, smartphone, hand
Voiceover: "Imagine going to bed with dirty clothes and waking up to a fresh, clean load. All without leaving the house. It's like having your own personal laundry assistant." 
Colors: Gray, Silver, Black, Dark_Gray, Dark_Brown

Scene 3: The scene shows a man and woman embracing while a washer and dryer are in the background
Emotions: love, happiness, togetherness
Tags: couple, hug, laundry, laundry room
Voiceover: "Never fight over the laundry room again."
Colors: Gray, Off_White, Silver, White,Black, Dark_Gray

Scene 4: The scene shows a logo of Maytag
Emotions: trust, confidence
Tags: logo, Maytag
Voiceover: "Get a Maytag and enjoy the luxury of time."
Colors: Off_White, Silver, White, Black, Dark_Gray, Gray
"
\end{lstlisting}

\begin{lstlisting}[caption={Advertisement generated by Henry-SEED for the Pruduential Financial Ad. Original Ad: \url{https://www.youtube.com/watch?v=JPfi0P-0nwU}},frame=single,breaklines=true,basicstyle=\tiny, label={lst:Prudential-financial-ad}]
Input Title:   "Financial Professional Tips: Planning for Your Own Care" for the brand Prudential Financial

Generated Ad: "The ad is shot in landscape orientation, at a low pace. 

Scene 1: The scene shows a woman looking thoughtfully at a laptop on her desk
Emotions: concern and contemplation
Tags: computer, laptop, office, woman
Colors: Gray, Dark_Blue

Scene 2: The scene shows the same woman now with a slight frown as she examines papers that appear to be medical bills
Emotions: worry and realization
Tags: bills, healthcare, paperwork, desk
Voiceover: "Long-Term Care is one of the most important issues in financial planning, but it's not always easy to talk about."
Colors: Gray, Dark_Blue

Scene 3: The scene shows the woman engaging in a conversation with a financial professional
Emotions: engagement and interest
Tags: consultation, financial advisor, discussion, office
Voiceover: "As a Financial Professional, I can help you evaluate your options and determine how much you may need to set aside each month."
Colors: Olive, Gray

Scene 4: The scene shows the woman in deep thought, looking out of a window while holding a pen and notepad, suggesting planning and reflection
Emotions: contemplation and seriousness
Tags: planning, reflection, window, notepad
Colors: Sky_Blue, Light_Gray

Scene 5: The scene transitions back to show the woman nodding in understanding as she listens to the financial professional, visibly gaining confidence
Emotions: comprehension and confidence
Tags: financial professional, office, learning, agreement
Voiceover: "Most people believe that they can rely on Medicare to pay for their long-term care expenses. But the truth is, Medicare typically only covers about 65%
Colors: Gray, Dark_Blue

Scene 6: The scene shows the woman now jotting down notes actively, with a look of determination on her face
Emotions: determination and proactivity
Tags: planning, notetaking, strategy, office
Voiceover: "So if you're a typical American, you'll need about $100,000 to cover the remaining costs of long-term care."
Colors: Gray, Dark_Blue

Scene 7: The scene shifts to the woman shaking hands with the financial professional, symbolizing a plan set in motion
Emotions: satisfaction and trust
Tags: handshake, agreement, financial planning, commitment
Colors: Gray, Dark_Blue

Scene 8: The scene cuts to the woman now at home, relaxed and smiling, speaking on the phone, perhaps sharing her relief and positive steps taken with a loved one
Emotions: relief and happiness
Tags: home, communication, personal life, reassurance
Voiceover: "Let's talk about a plan for your own care."
Colors: Cream, Soft_Green

Scene 9: The final scene zooms out to a serene image of the woman walking in a park, symbolizing peace of mind about her future
Emotions: contentment and peace
Tags: park, walking, future, peace of mind"
Colors: Sky_Blue, Green
\end{lstlisting}

\begin{lstlisting}[caption={Advertisement generated by Henry-SEED for an Adidas Ad. Original Ad: \url{https://www.youtube.com/watch?v=fujboh-W7Sk}},frame=single,breaklines=true,basicstyle=\tiny, label={lst:adidas}]
Input Title:  "I'M POSSIBLE | IMPOSSIBLE IS NOTHING | Adidas" for the brand Adidas


Generated Ad: "The ad is shot in landscape orientation, at a medium pace.

Scene 1: The scene shows a man facing a rock climbing wall, poised to start his ascent
Emotions: anticipation and focus
Tags: athletic, climb, climber, rock
Voiceover: "What is possible is up to you."
Colors: Gray, Dark_Brown

Scene 2: The scene shows a man as he begins to climb, gripping the rock tightly
Emotions: determination and strain
Tags: athletic, climb, climber, hand
Voiceover: "Don't let anyone tell you otherwise."
Colors: Dark_Gray, Dark_Brown

Scene 3: The scene shows the man midway up the rock wall, reaching for a distant handhold
Emotions: perseverance and challenge
Tags: athletic, climb, climber, hand
Colors: Black, Dark_Gray

Scene 4: The scene captures a moment where the man slips but regains his grip on the rock face, muscles tensing
Emotions: tension and danger
Tags: athletic, climb, climber, hand
Colors: Dark_Gray, Dark_Brown

Scene 5: The scene shows the man pausing to chalk his hands, his breath visible in the cool air
Emotions: calm and preparation
Tags: athletic, chalk, climb, climber
Colors: Gray, Dark_Brown

Scene 6: The scene shows the man navigating a particularly difficult part of the climb, his movements precise and deliberate
Emotions: concentration and determination
Tags: athletic, climb, climber, hand
Colors: Black, Dark_Gray

Scene 7: The scene shows the climber's face, sweat beading on his forehead, reflecting the intensity of his effort
Emotions: grit and resolve
Tags: athletic, climb, climber, determination
Colors: Dark_Gray, Dark_Brown

Scene 8: The scene shows the man reaching for the final ledge, the clear sky behind him signaling the nearing end of his ascent
Emotions: anticipation and effort
Tags: athletic, climb, climber, hand
Voiceover: "Impossible is not an obstacle."
Colors: Dark_Gray, Gray, Dark_Brown, Blue

Scene 9: The scene shows the man pulling himself onto the summit, his body silhouetted against the sky
Emotions: triumph and exhilaration
Tags: athletic, climb, climber, hand
Voiceover: "It's a chance to prove them wrong."
Colors: Dark_Gray, Dark_Brown, Blue

Scene 10: The scene shows the man standing on top of the rock with his arms raised in victory, overlooking the landscape below
Emotions: achievement and elation
Tags: athletic, arms up, climb, climber"
Colors: Dark_Gray, Dark_Brown
\end{lstlisting}

\begin{lstlisting}[caption={Advertisement generated by Henry-SEED for an Amazon ad. Original Ad: \url{https://www.youtube.com/watch?v=rcFTqs-zSH4}},frame=single,breaklines=true,basicstyle=\tiny,label={lst:amazon}]

Input Title:  "Alexa Auto, Listening to Audiobooks" for the brand Amazon

Generated Ad: "The ad is shot in landscape orientation, at a low pace.

Scene 1: The scene shows a car driving down a country road near a mountain house
Emotions: anticipation, freedom
Tags: blue, car, drive, house
Colors: White, Lavender, Dark_Blue,Turquoise, Black

Scene 2: The scene shows an image of a car approaching a large sign above the road, signaling the start of the highway
Emotions: curiosity, readiness
Tags: car, drive, road sign, highway
Voiceover: "You just pulled onto the highway"
Colors: White, Lavender,Turquoise, Gray

Scene 3: The scene shows an in-car view with a driver's hands on the steering wheel, a smartphone mounted on the dashboard displaying the Amazon Music app
Emotions: concentration, engagement
Tags: car interior, hands, smartphone, technology
Voiceover: " but you can't remember the name of that band that's been stuck in your head all morning."
Colors: White, Dark_Blue,Black, Gray

Scene 4: The scene shows the driver speaking to Alexa through the car's infotainment system, the display lighting up in response
Emotions: interaction, ease
Tags: car interior, voice command, infotainment system
Voiceover: "With Alexa and Amazon Music, you don't have to remember.Just ask, "Alexa, play the Beatles."
Colors: White, Lavender, Dark_Blue, Black

Scene 5: The scene shows the car passing through a landscape with rolling hills and the sun beginning to set in the background
Emotions: satisfaction, serenity
Tags: car, drive, hills, sunset
Colors: White, Lavender, Turquoise, Dark_Blue

Scene 6: The final scene shows the vehicle driving into the horizon, as the iconic music of the Beatles fills the car's interior
Emotions: fulfillment, joy
Tags: car, drive, horizon, music"
Colors: White, Lavender, Orange, Dark_Blue
\end{lstlisting}

\begin{lstlisting}[caption={Advertisement generated by Henry-SEED for an Gatorade ad. Original Ad: \url{https://www.youtube.com/watch?v=BKPQkjRF4yY}},frame=single,breaklines=true,basicstyle=\tiny, label={lst:gatorade}]
Input Title:  "Gatorade | Make Your Own Footsteps with Suni Lee" for the brand Gatorade

Generated Ad: "The ad is shot in landscape orientation, at a medium pace.

Scene 1: The scene shows Sunisa Manado standing in front of her house
Emotions: determination, brave, confident, persistence, courage
Tags: athletic, face, girl, muscle
Voiceover: "I'm Sunisa Manado, but you can call me Sunisa. And I'm going to show you how I stay motivated. Let's go! This is my neighborhood. I'm going to take you through my favorite spots. There's my house right there."
Colors: Dark_Brown, Black, Brown, Tan

Scene 2: The scene shows Sunisa Manado performing a handstand in the park
Emotions: achievement, determination, persistence, commitment, success
Tags: balance, gymnast, handstand, girl
Voiceover: "And this is the park where I get so much done. This is the park where I train."
Colors: Dark_Brown, Dark_Blue, Purple, Gray

Scene 3: The scene shows Sunisa Manado doing a flip on the balance beam
Emotions: brave, courage, determination, persistence, inspiration
Tags: gymnast, flip, beam, girl
Voiceover: "Being an athlete takes a lot of hard work and determination."
Colors: Dark_Brown, Dark_Blue, Purple, Gray

Scene 4: The scene shows Sunisa Manado in a powerful pose in her pink sports bra and leotard
Emotions: determination, brave, courage, persistence, inspiration
Tags: athletic, face, girl, gymnast
Colors: Dark_Brown, Dark_Blue, Purple, Gray

Scene 5: The scene shows Sunisa Manado lifting herself on the parallel bars
Emotions: achievement, persistence, determination, courage, commitment
Tags: gymnast, lift, bars, girl
Voiceover: "And being an athlete also means that you have to have good nutrition."
Colors: Dark_Brown, Dark_Blue, Purple, Gray

Scene 6: The scene shows Sunisa Manado practicing her routine on the floor of the park
Emotions: determination, persistence, inspiration, commitment, discipline
Tags: floor, gymnast, routine, girl
Colors: Dark_Brown, Dark_Blue, Purple, Gray

Scene 7: The scene shows Sunisa Manado spinning on the balance beam
Emotions: success, brave, determination, precision, focus
Tags: beam, spin, gymnast, girl
Colors: Dark_Brown, Dark_Blue, Purple, Gray

Scene 8: The scene shows Sunisa Manado holding a bottle of Gatorade, rehydrating after her practice
Emotions: refreshed, healthy, satisfaction, wellbeing, recovery
Tags: athlete, drink, Gatorade, health
Voiceover: "That's why I choose to fuel up with Gatorade. You need to have that energy, especially during my toughest training."
Colors: Black, Dark_Green, Dark_Gray, Gray

Scene 9: The scene shows Sunisa Manado smiling towards the camera with a sense of pride and accomplishment
Emotions: joy, confidence, pride, optimism, accomplishment
Tags: face, smile, girl, gymnast
Voiceover: "Gatorade helps replenish what I've lost and helps keep me going."
Colors: Black, Dark_Green, Dark_Gray, Gray

Scene 10: The scene shows Sunisa Manado walking away from the park, as the sun sets, symbolizing the end of a day's hard work
Emotions: contemplation, fulfillment, anticipation, resilience, hope
Tags: sunset, walk, reflection, girl
Voiceover: "I'm going to make my own footsteps. Follow your dreams, but don't forget to chase your greatness."
Colors: Black, Dark_Green, Dark_Gray, Gray"
\end{lstlisting}

\begin{lstlisting}[caption={Advertisement generated by Henry-SEED for a Rolex ad. Original Ad: \url{https://www.youtube.com/watch?v=vA-Lxmh1MdQ}},frame=single,breaklines=true,basicstyle=\tiny, label={lst:rolex}]
Input Title: "Rolesor A Meeting of Metals" for the brand Rolex

Generated Ad: "The ad is shot in landscape orientation, at a low pace."

Scene 1: The scene shows a close up of a watch that is on a wrist
Emotions: anticipation, luxury, elegance
Tags: dial, gold, watch
Colors: Dark_Brown, Brown, Black, Khaki

Scene 2: The scene shows a gold and silver watch sitting on a table, reflecting soft light
Emotions: sophistication, status, success
Tags: dial, gold, watch
Colors: Brown, Dark_Brown, Khaki, Black

Scene 3: The scene shows a person's hands elegantly wrapping the gold and silver watch in delicate gift paper
Emotions: care, generosity, surprise
Tags: wrapping, gift, luxury
Colors: Brown, Khaki, Gold, White

Scene 4: The scene shows a recipient's joyous reaction as they receive the watch as a gift at a celebratory gathering
Emotions: happiness, gratitude, celebration
Tags: emotion, celebration, gift
Colors: Gold, Silver, Cream, Warm hues, Soft lighting

Scene 5: The scene shows the watch being worn at an elegant event, the gold and silver glinting under ambient lighting
Emotions: pride, accomplishment, refinement
Tags: elegance, luxury, watch
Colors: Gold, Silver, Black, Soft lighting, Dark hues
\end{lstlisting}

\begin{lstlisting}[caption={Advertisement generated by Henry-SEED for an NVIDIA ad. Original Ad: \url{https://www.youtube.com/watch?v=HnaOhOJ6AcM}},frame=single,breaklines=true,basicstyle=\tiny,label={lst:nvidia}]
Input Title: "Pac Man 256 on NVIDIA SHIELD" for the brand Nvidia

Generated Ad: "The ad is shot in landscape orientation, at a medium pace. "

Scene 1: The scene shows a black and white logo.
Emotions: danger, tough
Tags: brand, logo
The text shown in the scene is 'SHIELD'
Colors: Black, White, Gray

Scene 2: The scene shows a black screen with a red light on it
Emotions: danger, tough
Tags: brand, logo, text
The text shown in the scene is 'SHIELD'
Colors: Black, White, Red

Scene 3: The scene shows a computer screen showing a game of Pac-Man
Emotions: nostalgia, fun
Tags: arcade, arcade machine, game, Pac-Man
The text shown in the scene is 'HIGH SCORE'
Colors: Cyan, Yellow, Gray, Dark Green, Red

Scene 4: The scene shows a video game character navigating a maze filled with obstacles
Emotions: excitement, challenge
Tags: arcade, character, game, maze
The text shown in the scene is 'LEVEL UP'
Colors: Cyan, Red, Pink, Orange, Gray

Scene 5: The scene shows a video game with a racing theme, cars zooming on the track
Emotions: thrill, speed
Tags: racing, cars, game, screenshot
The text shown in the scene is 'NEW RECORD'
Colors: Red, Yellow, Blue, Gray, Black

Scene 6: The scene shows a strategy video game with a detailed world map
Emotions: focus, strategy
Tags: strategy game, world map, empire, video game
The text shown in the scene is 'EXPAND YOUR EMPIRE'
Colors: Green, Brown, Blue, White, Tan

Scene 7: The scene shows a futuristic virtual reality game with immersive graphics
Emotions: amazement, innovation
Tags: VR, virtual reality, future, game
The text shown in the scene is 'ENTER NEW WORLDS'
Colors: Neon Blue, Neon Green, Black, Dark Blue

Scene 8: The scene shows a family gathered around playing a multiplayer video game, laughing and enjoying
Emotions: joy, togetherness
Tags: family, multiplayer, game, living room
The text shown in the scene is 'PLAY TOGETHER'
Colors: Multi-colored, Warm Hues

Scene 9: The scene transitions to a player wearing headphones, deeply concentrated in an esports competition
Emotions: determination, competition
Tags: esports, competition, gamer, video game
The text shown in the scene is 'RISE TO THE CHALLENGE'
Colors: Black, Neon, Dark, Spotlights

Scene 10: The scene closes with the NVIDIA SHIELD logo emerging brightly on the screen
Emotions: empowerment, assurance
Tags: brand, logo, closing
The text shown in the scene is 'NVIDIA. SHIELD - Game Your Way'
Colors: Red, White, Black
\end{lstlisting}

\begin{lstlisting}[caption={Advertisement generated by Henry-SEED for a Zillow ad. Original Ad: \url{https://www.youtube.com/watch?v=P6gydrvTkaA}},frame=single,breaklines=true,basicstyle=\tiny, label={lst:zillow}]
Input Title: "Find It, Tour It and Own It With Zillow (:06)" for the brand Zillow

Generated Ad: "The ad is shot in landscape orientation, at a medium pace.

Scene 1: The scene shows a person holding a cell phone with their hand.
Emotions: achievement, ambition
Tags: catch, smartphone, hand, person, phone, screen, text message, woman
Colors: Gray, Dark_Blue, Dark_Brown, Dark_Pink, Off_White

Scene 2: The scene shows a woman in a sweater looking at a cell phone.
Emotions: happy, smile
Tags: smartphone, earphone, girl, person, smile, woman
Colors: Gray, Dark_Brown, Dark_Pink, Dark_Blue

Scene 3: The scene shows a woman looking at a phone with a map on it.
Emotions: confident, smiling
Tags: catch, smartphone, earphone, girl, person, phone, stand, talk, woman
Colors: Gray, Dark_Brown, Dark_Pink, Dark_Blue

Scene 4: The scene shows a woman looking at a cell phone with a real estate app on it.
Emotions: togetherness, happiness
Tags: smartphone, earphone, girl, person, woman
Colors: Gray, Dark_Brown, Dark_Pink, Dark_Blue

Scene 5: The scene shows a woman looking at a cell phone with a real estate app displayed.
Emotions: happy, smiling
Tags: smartphone, earphone, girl, person, woman
Colors: Gray, Dark_Brown, Dark_Pink, Dark_Blue

Scene 6: The scene shows a woman using a cell phone to speak with an agent.
Emotions: confident, happy
Tags: smartphone, earphone, girl, person, talk, woman
Colors: Gray, Dark_Brown, Dark_Pink, Dark_Blue
Voiceover: "Getting the perfect home is a journey, so we help you find it, tour it, and own it."

Scene 7: The scene shows a woman completing a transaction on a cell phone.
Emotions: achievement, satisfied
Tags: smartphone, earphone, girl, person, woman
Colors: Gray, Dark_Brown, Dark_Pink, Dark_Blue"
Voiceover:  "For moving made simple, there's no place like Zillow"
\end{lstlisting}

\begin{lstlisting}[caption={Advertisement generated by Henry-SEED for a Kroger ad. Original Ad: \url{https://www.youtube.com/watch?v=SqwqI01q3fA}},frame=single,breaklines=true,basicstyle=\tiny, label={lst:kroger}]
Input Title: "How to Make Taco Seasoning | Kroger Recipes | Kroger" for the brand Kroger

Generated Ad: "The ad is shot in landscape orientation, at a low pace.  

Scene 1: The scene shows a person pouring chipotle adobo sauce into a glass jar.
Emotions: care, comfort
Tags: bottle, can, container, hand, food, person, jar, liquid, pepper, pour, red, sauce, tomato sauce, tin, tray, woman
Colors: Black, Gray, Dark_Brown, Maroon, Dark_Red
Voiceover: "The audio in the ad says Taco seasoning is one of those spices that everyone loves on their food. It's so delicious and it's so easy to make."

Scene 2: The scene shows a woman in a red sweater adding ground cumin to the mix.
Emotions: anticipation, focus
Tags: blender, container, food, hand, person, ingredient, measuring cup, spice, woman
Voiceover: "All you have to do is get your chopstick and add some of the garlic and some of the onion and some of the cumin"
Colors: Black, Gray, Dark_Brown, Maroon, Dark_Red

Scene 3: The scene shows the addition of chili powder to the seasoning mix.
Emotions: precision, satisfaction
Tags: blender, container, food, food processor, hand, person, ingredient, mixture, spice, woman
Voiceover: " and some of the dried chilies and some of the oregano and some of the salt"
Colors: Black, Gray, Dark_Brown, Dark_Pink, Maroon

Scene 4: The scene shows the woman sprinkling crushed red pepper into the mix.
Emotions: excitement, zeal
Tags: blender, container, food, food processor, hand, person, ingredient, measuring cup, mixture, spice, woman
Voiceover: "and some of the paprika and some of the pepper and just give it a little bit of a squeeze"
Colors: Black, Gray, Dark_Brown, Maroon, Dark_Red

Scene 5: The scene shows the woman measuring and adding ground oregano to the seasoning mix.
Emotions: happiness, creativity
Tags: blender, container, food, food processor, hand, person, ingredient, measuring cup, mixture, spice, woman
Colors: Black, Gray, Dark_Brown, Maroon, Dark_Red

Scene 6: The scene shows the final steps of making the taco seasoning mix, with the woman adding garlic powder and onion powder.
Emotions: accomplishment, pride
Tags: blender, container, food, food processor, hand, person, ingredient, mixture, spice, woman
Voiceover: "You're going to taste it, so don't be afraid to add more or less of any of these ingredients, depending on how spicy or how flavorful you want your taco seasoning to be."
Colors: Black, Gray, Dark_Brown, Maroon, Dark_Red"

\end{lstlisting}

\begin{lstlisting}[caption={Advertisement generated by Henry-SEED for a Publix ad. Original Ad: \url{https://www.youtube.com/watch?v=HH0VWvs-cFg}},frame=single,breaklines=true,basicstyle=\tiny, label={lst:publix}]
Input Title: "GreenWise Ice Cream Scoop from Publix" for the brand Publix Supermarkets

Generated Ad: "The ad is shot in landscape orientation, at a low pace. 

Scene 1: The scene shows a woman standing in front of a refrigerator.
Emotions: anticipation, satisfaction
Tags: beverage, bottle, catch, doorway, drink, fill, girl, home appliance, juice, laugh, pour, refrigerator, shirt, stand, woman
Voiceover: "Greenwise, the better energy choice from Publix."
Colors: Black, Gray, Dark_Gray, Dark_Brown, Silver

Scene 2: The scene shows a woman holding a roll of Greenwise ice cream in her hand.
Emotions: curiosity, interest
Tags: bottle, catch, hand, woman
Voiceover: " All the deliciousness of ice cream, a fraction of the calories. And now it comes in a roll."
Colors: Black, Gray, Dark_Gray, Dark_Brown

Scene 3: The scene shows the woman as she easily scoops the ice cream onto a plate, displaying the convenience of the new roll format.
Emotions: ease, delight
Tags: plate, roll, scoop, serve, woman
Voiceover: " So you can easily scoop and serve exactly what you need."
Colors: Black, Gray, Dark_Gray, Dark_Brown

Scene 4: The scene shows the woman enjoying a bite of the Greenwise ice cream, her expression reflecting the product's deliciousness.
Emotions: enjoyment, satisfaction
Tags: bite, enjoyment, ice cream, satisfaction, taste, woman
Voiceover: "It's a win-win. Learn more at publix.com."
Colors: Black, Gray, Dark_Gray, Dark_Brown"

\end{lstlisting}

\begin{figure*}[]
    \centering
    \includegraphics[width=0.99\textwidth]{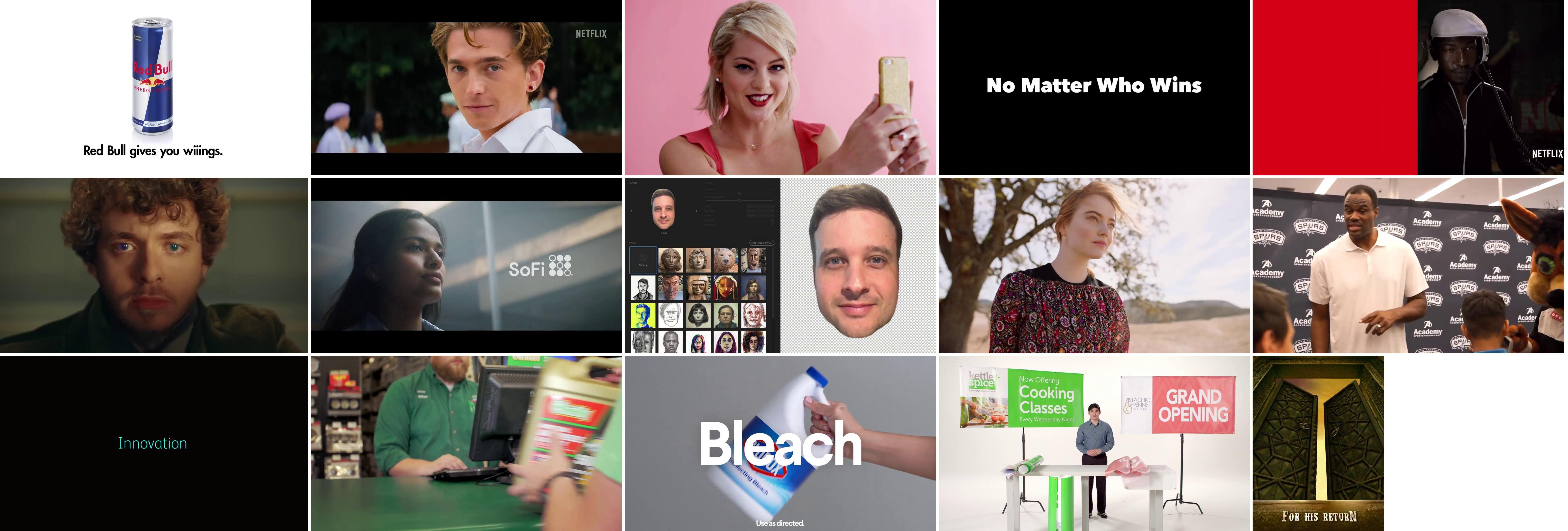}
    \label{fig:memorabilitykeyframes}
    \centering
    \includegraphics[width=0.99\textwidth]{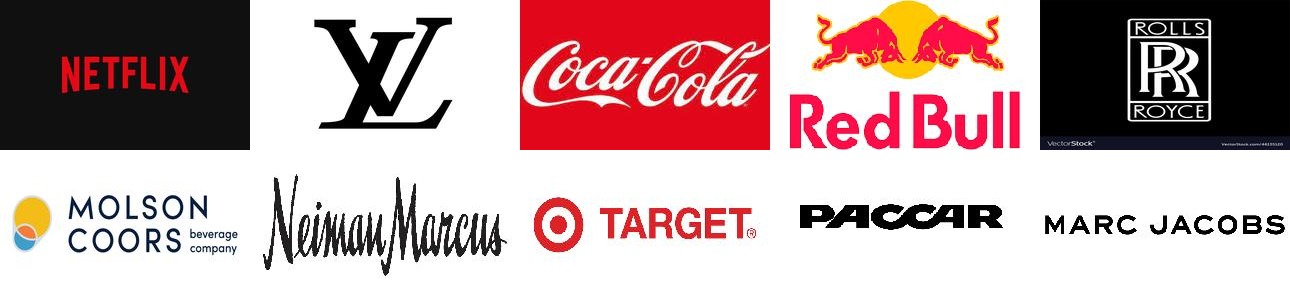}
    \caption{The top three rows show the keyframes from videos in our dataset, LAMBDA, arranged from most to least memorable. The bottom two rows show brands arranged from the most memorable brands to the least. 
    }
    \label{fig:Memorable brands}
\end{figure*}

\begin{figure}
    \centering
    \includegraphics[width=0.48\textwidth]{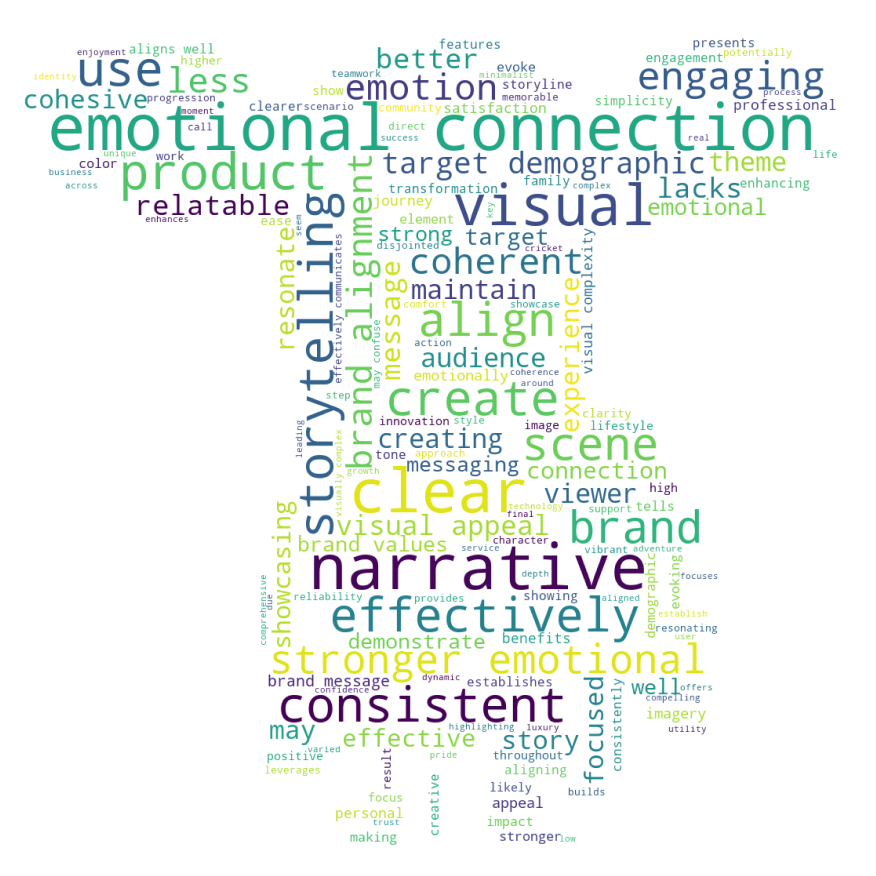}
    \caption{Word Cloud for the GPT-4 reasoning on the 75/88 generations where it rates Henry-SEED Generated Ads to be better than the Original.}
    \label{fig:gpt4-henry-win-wordcloud}
\end{figure}

\begin{figure}[]
    \centering
    \includegraphics[width=0.48\textwidth]{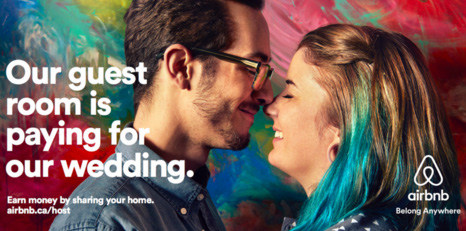}
    \caption{Airbnb advertisement showing the visual concepts of two adults, and the text ``Our guest room is paying for our wedding''. ``World knowledge'' captured by LLMs helps identify the two adults as partners, and helps relate the text with the two adults and the Airbnb logo to infer what the ad is talking about.}
    \label{fig:airbnb-ad}
\end{figure}

\FloatBarrier
\section{Questionnaire to Gather Human Preferences over Generated Ads}
\label{sec:Questionnaire to Gather Human Preferences over Generated Ads}
Below is the web-based form used to annotate the human preferences between the generated and original ad stories. Participants for this task were working professionals in the software, marketing, advertising, and creative industries. Participation was voluntary, and participants were invited to judge the efficacy of generated advertisements. Participants had a general interest in the creative and advertising industries and generative technologies; therefore, they were not interested in getting paid but rather in seeing and trying out the generative technology stack. We have a roughly 65-35 distribution of males to females with the age range between 22-50.

\begin{lstlisting}[caption={},frame=single,breaklines=true,basicstyle=\tiny, label={lst:form}]

Instructions:

Shown next are 10 pairs of advertisements. Determine which ad within each pair is more effective based on the title, brand, and scene-by-scene descriptions provided. You will also be expected to provide reasons for your choice wherever asked.


Question 1
Choose the advertisement you find more effective. Also provide reasons for your choice.

Title: Bike to Work Day at NVIDIA
Brand: Nvidia
Nvidia is a technology company focusing on graphics processing units (GPUs) for gaming, professional visualization, data centers, and automotive markets, driving innovation in visual computing.

Advertisement A:

The ad is shot in landscape orientation, at a medium pace. The audio in the ad is silent.
Scene 1: The scene shows the camera takes a photo from the inside of the person on the bicycle
Colors: White, Dark_Pink, Olive, Gray, Pink, Dark_Brown
Emotions: danger, dangerous, warning
Tags: attach, bicycle, catch, smartphone

Scene 2: The scene shows the person riding a bicycle down the road
Colors: White, Dark_Gray, Mud_Green, Olive, Gray
Emotions: danger, quiet
Tags: bicycle, path, grass, motorbike
The text shown in the scene is 'NVIDIA' 

Scene 3: The scene shows a man on a bike taking a ride
Colors: Off_White, Dark_Gray, Silver, Black, Gray
Emotions: danger, exciting, fun
Tags: bicycle, biker, bridge, hand
The text shown in the scene is 'DVIDIA' 

Scene 4: The scene shows a bike rider going under a bridge under a road
Colors: Dark_Gray, Silver, Light_Green, Green, Olive, Gray, Bright_Green
Emotions: danger, dangerous, funny
Tags: bridge, car, curve, highway
The text shown in the scene is 'NVIDIA' 

Scene 5: The scene shows a man riding a bicycle down a tree lined street
Colors: White, Dark_Gray, Mud_Green, Dark_Pink, Olive, Black, Gray
Emotions: thrill, adventure, romantic
Tags: bicycle, biker, hand, person
The text shown in the scene is 'NVIDIA' 

Scene 6: The scene shows a man riding on a bicycle down the street
Colors: Emerald, Dark_Gray, Silver, Light_Green, Olive, Gray
Emotions: funky, enjoyable
Tags: bicycle, hand, person, man
The text shown in the scene is 'NVIDIA' 

Scene 7: The scene shows a closeup of someone riding a bicycle down a road
Colors: White, Dark_Gray, Silver, Dark_Pink, Olive, Gray
Emotions: danger, majestic
Tags: bicycle, bicycle helmet, biker, hand
The text shown in the scene is 'NVIDIA' 

Scene 8: The scene shows a person is riding a bike on the side of the road
Colors: White, Dark_Gray, Mud_Green, Olive, Gray, Lavender
Emotions: enjoy, enjoyment
Tags: car, person, man, motorcycle
The text shown in the scene is 'NVIDIA' 

Scene 9: The scene shows someone riding a bike in front of a small city
Colors: White, Dark_Gray, Olive, Black, Gray
Emotions: funky
Tags: bicycle, biker, bin, car
The text shown in the scene is 'NVIDIA' 

Scene 10: The scene shows a cyclist riding his bike on a gravel road
Colors: White, Brown, Mud_Green, Olive, Gray, Dark_Brown, Cyan
Emotions: recreational, adventure
Tags: bicycle, biker, hand, person

Advertisement B:

The ad is shot in landscape orientation, at a low pace. The audio in the ad is silent.

Scene 1: The scene shows a man wearing a hard hat holding a bike helmet
Colors: Dark_Gray, Brown, Mud_Green, Cream, Olive, Black, Dark_Brown
Emotions: protective, protective
Tags: building, construction worker, hat, jumpsuit

Scene 2: The scene shows a man riding a bike on a path near a creek
Colors: Emerald, Dark_Gray, Mud_Green, Olive, Black, Dark_Brown
Emotions: recreational, relaxation
Tags: bicycle, bicycle helmet, biker, path

Scene 3: The scene shows a man holding a bike up while standing in front of a building
Colors: Dark_Gray, Brown, Mud_Green, Cream, Olive, Black, Dark_Brown
Emotions: pride, achievement
Tags: building, professional, hat, bicyclist

Scene 4: The scene shows a man riding a bike down a street with trees lining the road
Colors: Brown, Cream, Green, Olive, Dark_Brown
Emotions: cheery, freedom
Tags: bicycle, bicycle helmet, biker, man

Scene 5: The scene shows a man riding a bike down a street in front of a house
Colors: Dark_Gray, Mud_Green, Olive, Black, Dark_Brown
Emotions: cheery
Tags: bicycle, bicycle helmet, biker, car

Scene 6: The scene shows a closeup of the man's face as he adjusts his bike helmet, showcasing determination
Colors: Cream, Olive, Black, Gray, Dark_Brown
Emotions: determined, prepared
Tags: man, helmet, focus, detail

Scene 7: The scene shows the man holding his bike next to other cyclists at a traffic light, promoting community and camaraderie
Colors: Mud_Green, Cream, Olive, Dark_Brown
Emotions: community, anticipation
Tags: cyclists, traffic light, group, waiting

Scene 8: The scene shows the man arriving at work, parking his bike in a bike rack
Colors: Mud_Green, Cream, Olive
Emotions: satisfaction, accomplishment
Tags: office building, bike rack, arrival, work

Scene 9: The scene shows the man walking into the building, greeting colleagues who are also carrying bike helmets
Colors: White, Cream, Olive, Black, Gray
Emotions: friendly, inclusive
Tags: workplace, colleagues, greeting, professional attire

Scene 10: The scene shows the man at his workstation with a helmet on his desk, looking out the window at the sunny day, hinting at the ride home
Colors: White, Cream, Olive, Gray
Emotions: thoughtful, accomplished
Tags: office, workstation, helmet, window


Select preferred advertisement:
Option 1: A
Option 2: B
Option 3: Both are equally effective


Give reasons for your choice:
______________________________________
\end{lstlisting}
\subsection{Expert Feedback Collected For Generated Ads}
\label{sec:Expert Feedback Collected For Generated Ads}
\begin{enumerate}
    \item Feedback for ad generation for the Maytag Ad shown in Fig~\ref{fig:maytag-generated-ad}
    \begin{enumerate}
        \item \textbf{Expert 1}: "I appreciate the prominent use of the logo in the advertisement. Its placement towards the end, accompanied by a compelling slogan, is in alignment with the brand's advertising strategy."
        \item \textbf{Expert 2}: "In my opinion, the color scheme of the advertisement is stunning. It complements the tone of the advertisement exceptionally well."
        \item \textbf{Expert 3}: "The emotional portrayal in scene 2 could be enhanced. I anticipated a sense of 'recreation' and 'relaxation' to be more effectively conveyed."
    \end{enumerate}
    \item Feedback for ad generation for the New York Times Ad shown in Fig~\ref{fig:adgen-NYT}
    \begin{enumerate}
        \item \textbf{Expert 1}: "One noteworthy aspect in the generated ad description is the concept of 'blocking.' In the ad, the main actor is depicted moving and protesting against various backdrops, including a static background and a subtly shifting frame. This technique is reminiscent of the famous concept utilized in cinematography. While this is not in reflected in the image, I will attribute it to the image generation and not the description generation."
        \item \textbf{Expert 2}: "I like the generated voiceover a lot in terms of story, but I find it hard to fit over the scenes, perhaps this is because the generations dont incorporate transitions/animations."
        \item \textbf{Expert 3}: "I find the overall generated story exceptional in terms of its storytelling in a few ways. 1. The flow of the generated ad, A woman exploring nightlife, protesting, achieving, and nonetheless standing defiant. 2. The slogans are great. 3. The changing head tilt of the woman from sideways to center is a very precise details cinematographer use to paint an overall story or emotion." 
    \end{enumerate}
    \item Feedback for ad generation for the Brainly Ad shown in Fig~\ref{fig:brainly-generated-ad}
    \begin{enumerate}
        \item \textbf{Expert 1}: "I find the overall story formulation to be decent. It portrays kids encountering challenges in solo learning, showcasing easy accessibility and a gradual improvement in confidence and engagement throughout the story. I would still prefer a scene where the UI of the app is somehow shown to the user.\footnote{The generated description of the ad actually shows the student interacting with a visible UI that the image generation model could not respect properly}"
        \item \textbf{Expert 2}: "I like the use of animated scenes, but I find the incorporation of different main characters slightly jarring. Either they should have been in a common scene, or the main character should not change with every scene. The standout feature of the ad is the utilization of color themes and their harmonization with the emotional tone of each scene."
        \item \textbf{Expert 3}: "Having created Ed-Tech advertisements, I find the storytelling to be excellent. This ad is very persuasive, although it lacks novelty, I still find it to be effective."
    \end{enumerate}
\end{enumerate}

\section{Perplexity evaluation}
\label{sec:Perplexity evaluation}
A common approach to measuring language modeling performance on some data distribution $D$ is to measure \textit{perplexity}, which is defined as the exponential of the average negative loglikelihood per token \cite{perplexityjelinek, brown1992estimate,biderman2024lessons}, that is:
\begin{equation}
PPL = \exp\left(\frac{-1}{\sum_{j=1}^{|D|}N_j}\sum_{j=1}^{|D|}\sum_{i=1}^{N_j}\log P(y_{j_i}|y_{j_1}, \dots, y_{j_{i-1}})\right),
\end{equation}\label{eqn:ppl} where $|D|$ is the number of documents in the dataset, $y_j$ is the $j$-th document in $D$, $N_j$ is the total number of tokens in $y_j$, and $y_{j_i}$ represents the $i$-th token of $y_j$.

To calculate perplexity on a selected dataset $D$, each dataset document $y$ is tokenized and fed into a language model (following the procedure described below) via computing $\log P(y|x)$, where $x$ is set to either the empty string or a beginning-of-text token. Thus, given $\log P(y)$, for each document $y \in D$ we can sum up the per-document loglikelihoods and divide by the number of total dataset tokens.

Given our language model, we aim to compute the conditional (log) probability (or ``loglikelihood'') of a target string $y$ conditioned on input $x$, denoted as $\log P(y|x)$. This can be performed in a single LM call.

Let $x = x_0, x_1, ..., x_{n-1}$ be an input sequence of $n$ tokens and $y = y_{0}, y_{1}, ..., y_{m-1}$ be the target sequence of $m$ tokens, where $x_i$ and $y_i$ represent individual tokens. To compute $\log P(y|x)$, we follow these steps:

\begin{enumerate}
    \item Concatenate $x$ and $y$ to form a new sequence, but discard the final token $y_{m-1}$. The resulting sequence is $x_0, x_1, ..., x_{n-1}, y_{0}, y_{1}, ..., y_{m-2}$.
    \item Pass this concatenated sequence through the language model to obtain logits $l$ of shape $(n + m - 1, |V|)$, where $|V|$ is the size of the vocabulary. The last $m$ positions in these logits correspond to the predicted probability distributions for the target tokens $y_0$ to $y_{m-1}$, conditioned on the input $x$ and the preceding target tokens.
    \item Apply a log-softmax function to the last $m$ logits to obtain log probabilities for the completion tokens only.
    \item Calculate the conditional loglikelihood of the target string $y$ given the input $x$ by summing the log probabilities of each target token:
    \begin{equation} \label{eqn:logp}
    \log P(y|x) = \sum_{i=0}^{m-1} \log p(y_{i}|x, y_0, ..., y_{i-1}) = \sum_{i=0}^{m-1} l(n +i, y_i),
    \end{equation} where $\log p(y_i|x, y_0, ..., y_{i-1})$ is the log probability of the $i$-th target token conditioned on the full input $x$ and the preceding target tokens. (and where $x, y_0,... y_{-1}$ denotes conditioning on only $x$.)
\end{enumerate}

We follow the above procedure to calculate perplexity over three equally divided parts of the dataset, \textit{i.e.}, 33-percentile cuts where samples are ranked as per their memorability values. The lower the perplexity of an LLM over a category of samples, the better it is at generating those samples. Therefore, for example, if an LLM has a lower perplexity over high memorable samples, it is easier for it to generate highly memorable samples than lower memorable ones.

\section{Annotation Protocol and Participant Details for the LTM Study}
\begin{figure*}[!h]
    \centering
    \includegraphics[width=\textwidth]{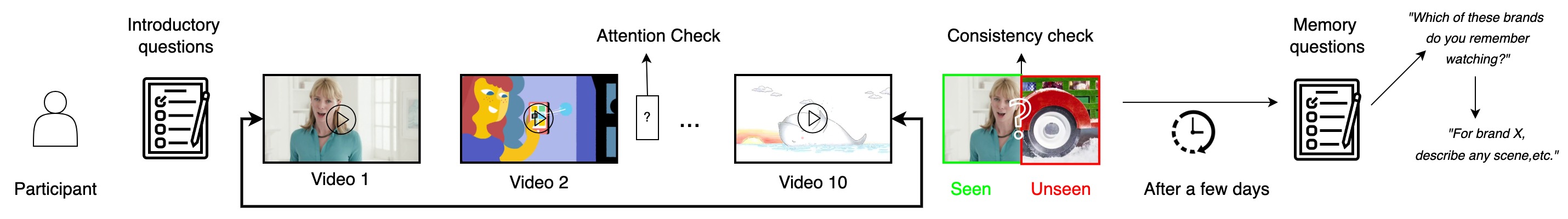}
    \caption{The study protocol we followed for our long term memorability human study. All the previous works follow a game-like annotation protocol, where the study participants compete with each other to get best memorability scores and a participant is excluded from the study if their annotations fall below a certain threshold. We follow a more natural way in which participants fill an initial questionnaire, then watch 10 ads with attention checks on day 1 and in subsequent days, receive a form asking them to fill in what do they remember seeing. Further, using Stable Diffusion, we also ask them to recreate the scenes they remember.}
    \label{fig:study protocol}
\end{figure*}

Figure~\ref{fig:study protocol} shows a visualization of the annotation protocol we followed.

The participants in the study were students who were offered optional course credit and freebies like eatables and a chance to see research and know their memorability scores. The participation was voluntary. The students were shown a protocol of the study and were required to sign the IRB approval, which was prominently displayed. The approval contained details about what kind of data was being collected and how the data would be used. The data collection protocol was approved by the IRB of the participating institution. The aggregate statistics were reported to each candidate after completing the study. Three emails were sent to take-home participants; if they didn't reply within the given time frame, their data was discarded from the experiment.

The participants were primarily graduate and undergraduate students. The participants are from two universities spread across two locations in India. The participants are bilingual and speak a variety of languages, including English. The age range is from 16 to 35 years, and all genders/sexes are encouraged. We saw a roughly 30-70 distribution of females to males.

\subsection{Memorability Questionnaire}
\label{sec:Memorability Questionnaire}
This section contains the questions we asked before the study, the attention check questions that were asked during the study, and finally, the recognition questions to check which brands were remembered.

\subsubsection{Introductory Questionnaire (to be filled before the study starts)}

\begin{enumerate}
    \item I remember seeing ads for the following brands this year:
    \begin{itemize}
        \item List 15 randomly selected from the list of brands that we have
    \end{itemize}
    
    \item I remember using products of the following brands this year:
    \begin{itemize}
        \item List 15 randomly selected from the list of brands that we have (non-intersecting list from above)
    \end{itemize}
    
    \item Have you installed any Ad Blocking software in your browser(s)?
    \begin{itemize}
        \item[a.] Yes
        \item[b.] No
    \end{itemize}
    
    \item Do you use a Youtube subscription?
    \begin{itemize}
        \item[a.] Yes
        \item[b.] No
    \end{itemize}
    
    \item Approximately how much percentage of time do you spend on Youtube mobile vs Youtube web?
    \begin{itemize}
        \item $<$10\% on mobile
        \item $>$10\% but $<$30\% on mobile
        \item $>$30\% but $<$70\% on mobile
        \item $>$70\% on mobile
    \end{itemize}
    
    \item How do you apprise yourself of the latest products and brands? (Multi correct)
    \begin{itemize}
        \item Primarily friends and family
        \item Amazon, Flipkart or any other e-commerce stores
        \item Television and OTT Platform Ads (like Youtube, Netflix, Hotstar, etc)
        \item Email Ads
        \item Store Visits
        \item Website Ads
        \item I primarily search for products
    \end{itemize}
    \end{enumerate}

\subsubsection{Checks (to be answered during the experiment)}
    \begin{enumerate}
        \item \textbf{Attention check}: A factual question like, What is the capital of India? (Asked randomly between videos, needs to be answered in $<$10s)
        \begin{itemize}
            \item[a.] Kanpur
            \item[b.] Delhi
            \item[c.] Goa
            \item[d.] Mumbai
        \end{itemize}
        \item \textbf{Consistency Check}:  Do you remember watching this video in this experiment (Asked after showing the 11th video)
    \begin{itemize}
        \item[a.] Yes
        \item[b.] No
    \end{itemize}
    \end{enumerate}

\subsubsection{Recognition Questions (asked after a few days after watching the videos)}
\begin{enumerate}
    \item In the study, I remember seeing Ads of the following brands:
    \begin{itemize}
        \item (Randomly selected list of 20 brands which contains the brands shown to the participant)
        \item \{For each brand in the list which the participant has selected\}
    \end{itemize}
    
    \item Brand: X (already filled in)
    \begin{itemize}
        \item For the \{brand\} ad, I remember seeing the following (Write Scene Descriptions, feel free to write any scenes, music, characters, emotions, objects you remember seeing):
    \end{itemize}
\end{enumerate}

\section{Collection of all the Prompts used in the Paper}

\subsection{GPT-4 Prompts}
\begin{lstlisting}[caption={GPT-4 Prompt to calculate preference between Real Ad (A) and Generated Ad (B)},frame=single,breaklines=true,basicstyle=\tiny, label={lst:ad-quality-preference-prompt}]
As a seasoned marketer, evaluate the effectiveness of the following two ads using a comprehensive set of metrics:

Creativity and Innovation: Originality and uniqueness in conveying the message. Use of unexpected ideas or elements that grab viewers' attention.

Emotional Connection: Ability to evoke strong, relevant emotions in the target audience. Establishing a connection between the brand and the viewers' emotions.

Storytelling: Crafting a compelling narrative that engages and retains the audience. Creating a memorable experience through a coherent and impactful story.

Visual Appeal: Use of strong visual elements, such as striking visuals, colors, and graphics. Ensuring that the visual elements align with the overall message and brand image.

Brand Alignment: How well the ad aligns with the values, mission, and personality of the brand. Consistency with the brand's visual identity, tone, and messaging. The ad's ability to leave a lasting impression on viewers regarding the brand. Incorporating brand elements that make it easy for the audience to remember and recognize.

Target Demographics: Relevance of the ad content and message to the target audience. Appropriateness of visuals, language, and themes for the specific demographic group.

Based on these criteria, analyze and determine which of the two ads is more effective. I will provide you with the Voiceover, followed by their scene-by-scene descriptions, including the emotions shown in the scene, the text, objects, colors, and style of the image.

Ad (A): {Verbalization for Ad (A)}

Ad (B): {Verbalization for Ad (B)}

Give me your answer in a json format, with the following keys:
- ad_a_score: Score between 0 and 10 for Ad A
- ad_b_score Score between 0 and 10 for Ad B
- winner The winner of the two ads
- reason line separated Reasons for the winner in not more than 3 lines
\end{lstlisting}

\begin{lstlisting}[caption={GPT-4 Prompt to measure consistency of an Ad},frame=single,breaklines=true,basicstyle=\tiny, label={lst:ad-quality-consistency-prompt}]
You are now a seasoned marketer that judges the consistency of an advertisement well. The consistency of an Ad can be determied by a few metrics (in no particular order) such as:
1. Does the voiceover match with the Scenes in the Ad?
2. Do the scene description make a good story?
3. Are the emotions depicted in the scenes consistent with the overall ad?
4. Does the ad represent the product and the brand well?

Rate the consistency of the following ad out of 10. Give me the rating only and nothing else, or you will be penalized.
{Advertisement Description}
\end{lstlisting}

\begin{lstlisting}[caption={GPT-4 Prompt to generate ad verbalization with In-Context-Learning (ICL)},frame=single,breaklines=true,basicstyle=\tiny, label={lst:ad-gen-prompt}]
You are now a seasoned marketer that creates memorable ads given its duration, brand and title.
Your output should follow the writing style of the input exactly. For example, each scene should look like:
The scene shows {}. The foreground colors of the scene are {}, and the background colors are {}. The dominant tone of the scene is {}. The photography style of the scene is {}. The scene has {} visual complexity. The emotions shown in the scene are {}.  This scene is categorized by the tags {}. 
You are only supposed to fill in the {}

Generate the detailed description of a {DURATION_AD1} second memorable advertisement titled "{TITLE_AD1}" for the brand {BRAND_AD1}
Generate the detailed description of a {DURATION_AD2} second memorable advertisement titled "{TITLE_AD2}" for the brand {BRAND_AD2}
...
Generate the detailed description of a {DURATION_AD5} second memorable advertisement titled "{TITLE_AD5}" for the brand {BRAND_AD5}
Generate the detailed description of a {DURATION_TARGET} second memorable advertisement titled "{TITLE_TARGET}" for the brand {BRAND_TARGET}
\end{lstlisting}

\subsection{Henry Prompts}
\label{sec:Henry-prompts}
Given below are the verbalization templates we use to teach Henry and Henry-SEED behavior simulation and content simulation tasks:

\begin{lstlisting}[caption={Verbalization pattern to predict memorability given advertisement. The same template is used to prompt GPT-3.5, GPT-4, Henry, Henry-Oracle, and Henry-SEED. Note that video tokens are optional.},frame=single,breaklines=true,basicstyle=\small\ttfamily,label={listing:memorability-prediction-verbalization-format}]
Students are shown ads and their memorability is tested after 1 to 3 days. For the given ad: 
<video> .. </video> 
They watch a 15 second advertisement for Chanel. 
The title of the advertisement is " Comes in Red for a Limited Edition CHANEL Fragrance". 
The ad is shot in landscape orientation, at a medium pace.
The audio in the ad says: Number 5. Limited Edition. Chanel. 
Following are the descriptions of each scene: 
    Scene 1: 
        The scene shows a red bottle of perfume that is on a dark surface.
        The foreground colors of the scene are Black, and the background colors are Dark_Brown,Maroon,Black,Gray.
        The dominant tone of the scene is neutral.
        The photography style of the scene is product. 
        The scene has Low visual complexity.
        The emotions shown in the scene are gift, romantic, celebration. 
        This scene is categorized by the tags bottle, man, perfume, red, woman.
        The text shown in the scene is 'N5', 'CHANEL', 'PARIS', 'PARFUM' 
        ....
What would be the memorability score of this video?

Output: 71
\end{lstlisting}

\begin{lstlisting}[caption={Henry Prompt to generate ad verbalization used to train and evaluate Henry-SEED},frame=single,breaklines=true,basicstyle=\tiny,label={listing:advertisement-generation-prompt-Henry-SEED}]
Generate the detailed description of a {DURATION_TARGET} second memorable advertisement titled "{TITLE_TARGET}" for the brand {BRAND_TARGET}
\end{lstlisting}

\subsection{Mistral prompt for filtering marketing ads}
\label{lst:msitral-filter-prompt}
\begin{lstlisting}[caption={Mistral Prompt for Ad Filtering},label={listing:Mistral Prompt for Ad Filtering},frame=single,breaklines=true,basicstyle=\tiny,]
"Based on the topic_tags_vocab = {'politics': 'The art and science of governing societies and making decisions that affect collective interests.', 'marketing': 'The process of promoting, selling, and distributing products or services to consumers, often involving market research, advertising, and branding strategies.'} provided, please identify the top most relevant topic tag from the topic_tags_vocab keys that represent the following advertisement based on content and page_name. Please use only the most relevant tag and make sure to choose from provided topic tags only. Do not include any other tags not mentioned in the prompt.Answer with the most relevant topic tag only. The advertisement is posted by the page Donald J. Trump and has the following content : ['President Trump is coming to town! Get your free tickets now >>>']. Answer in only politics or marketing."

cleaned_text = "The advertisement is posted by the page {page_name} and has the following content : {page_content}"
\end{lstlisting}

\section{Computing Infrastructure and Hyperparameters}
\subsection{Modeling Memorability}
\label{sec:experimental-details}
All the experiments were conducted on 8x40 A100 instances. All experiments were performed leveraging DeepSpeed ZeRO stage-3 with cpu offload \cite{ren2021zero,rasley2020deepspeed,rajbhandari2020zero} and Flash-attention \cite{dao2022flashattention} with gradient-checkpointing \cite{chen2016training} at bf16 precision. We use AdamW as the optimizer (with fused gelu), the learning rate was kept 2e-5 for all experiments. The maximum context length for image-only datasets is 500, including public video datasets is 800 and including our dataset is 2048. The corresponding batch sizes are 32,16,8. The gradient accumulation is set to 1 and weight decay is disabled. The warmup steps are set to 20 and residual dropout was kept at 0.25. We train all models for two epochs, but use the checkpoint with best validation spearman correlation.

For all experiments, where we combine datasets, we use a custom sampler to account for dataset imbalance, that ensures a maximum proportion of the dataset in an epoch, here are the maximum proportions. For validation we take 5\% of each dataset.
We use the provided test splits for public datasets and we use a 15\% test split for our dataset

\subsubsection{Images}
\begin{enumerate}
    \item \textbf{Lamem} 50\%
    \item \textbf{Memcat} 100\%
    \item \textbf{SUN} 100\%
\end{enumerate}

\subsubsection{Videos}
\begin{enumerate}
    \item \textbf{VideoMem} 75\%
    \item \textbf{Memento} 75\%
    \item \textbf{AdsData} 100\%
    \item \textbf{MediaEval} 100\%
\end{enumerate}

\subsection{Generating Memorable Ads}
All the experiments were conducted on 8x80 A100 instances. All experiments were performed leveraging DeepSpeed ZeRO stage-2, Flash Attention and Gradient-Checkpointing.
$\alpha=0.001$, awac\_scale$=1$, $\gamma=0.99$, $\beta=0$ cql\_scale$=0.1$
\subsubsection{Inference hyperparameters}
$\beta=4$, temperature$=0.8$, steps\_for\_target\_sync 10, 
$\tau=0.7$, two\_qs: True, lr=1e-5

\section{License and Terms of Release}
LAMBDA and UltraLAMBDA are sourced from brand videos from YouTube, Facebook Ads, and CommonCrawl. The dataset annotations and video links contained in LAMBDA and UltraLAMBDA will be released under CC BY-NC 4.0 license. The videos themselves are released as per their creators' licenses. The videos or the released data do not contain or disclose any identities of their annotators or any specific persons. Since it is handcrafted, LAMBDA makes sure that none of the videos are offensive; UltraLAMBDA being sourced from the internet is noisier. While the videos themselves originate from brands, the content of some brands may seem offensive to certain people. 

We used Llama, GMHRA, ViT, EVA-CLIP, and Qformer models in accordance with their licenses to train Henry.

\section{Limitations and Potential Risks}
In this paper, we try to fill a gap in the existing literature about long-term memorability modeling and datasets. Therefore, we conduct the first study for that purpose. While doing that, we have made initial efforts starting with the English language advertisements. Future work would be needed to address other languages. Further, given the limitations of the study, we conducted it in an academic environment with a student population consisting of undergraduate and graduate student volunteers. We will expand the scope to a wider audience in the future work. We trained a model, Henry, on the collected dataset, showing good performance on all literature datasets. However, since the literature datasets are all English-based and deal with a majorly uniform population, the training will be scaled to more languages and population types in future work. We also observed a decrease in performance for brands not seen during the training and for videos with longer verbalizations exceeding 1500 tokens. Additionally, the model exhibits a slight inaccuracy when advertisements have significant musical content. In our opinion, the model does not pose any potential risk or harm besides the limitations mentioned here. We also conduct a review of the generated ads through experts and non-expert annotators. Both experts and non-expert annotators preferred Henry-SEED generated ads 3/5 times.

\end{document}